\documentclass[10pt,twocolumn,letterpaper]{article}

\usepackage{microtype}
\usepackage[pagenumbers]{cvpr} %

\usepackage[T1]{fontenc}
\definecolor{commentpurple}{rgb}{0.7,0.2,0.9}

\definecolor{newtextblue}{rgb}{0.05,0.35,0.75}
\definecolor{citeorange}{rgb}{0.85,0.4,0.0}

\newcommand{\first}[1]{{\colorbox{red!40}{#1}}}
\newcommand{\second}[1]{{\colorbox{orange!50}{#1}}}
\newcommand{\third}[1]{{\colorbox{yellow!50}{#1}}}
\definecolor{cvprblue}{rgb}{0.21,0.49,0.74}
\usepackage[pagebackref,breaklinks,colorlinks,allcolors=cvprblue]{hyperref}
\usepackage{xcolor}
\definecolor{suppImprove}{RGB}{0,110,70}
\definecolor{suppRegress}{RGB}{180,45,45}
\newcommand{\suppgain}[1]{\textcolor{suppImprove}{#1}}
\newcommand{\suppdrop}[1]{\textcolor{suppRegress}{#1}}

\definecolor{cvprblue}{rgb}{0.21,0.49,0.74}
\usepackage[pagebackref,breaklinks,colorlinks,allcolors=cvprblue]{hyperref}

\def\confName{CVPR}
\def\confYear{2027}

\title{Fix4D: Repurposing Video Generative Priors for 4D Tracking Refinement}

\author{Rui Wang \quad Haofei Xu \quad Quentin Lohmeyer\quad Siyu Tang \quad Mirko Meboldt\vspace{0.3em} \\
{\normalsize ETH Z\"urich} \\
{\tt\small \{ruiwang46, qlohmeye, meboldt\}@ethz.ch} \quad\quad {\tt\small \{haofei.xu, siyu.tang\}@inf.ethz.ch}\\
{\normalsize \url{https://batfacewayne.github.io/Fix4D.io/}}
\vspace{-5mm}
}
\begin{document}

\twocolumn[{
\renewcommand\twocolumn[1][]{#1}%
\maketitle
  \begin{center}
    \includegraphics[width=1\textwidth]{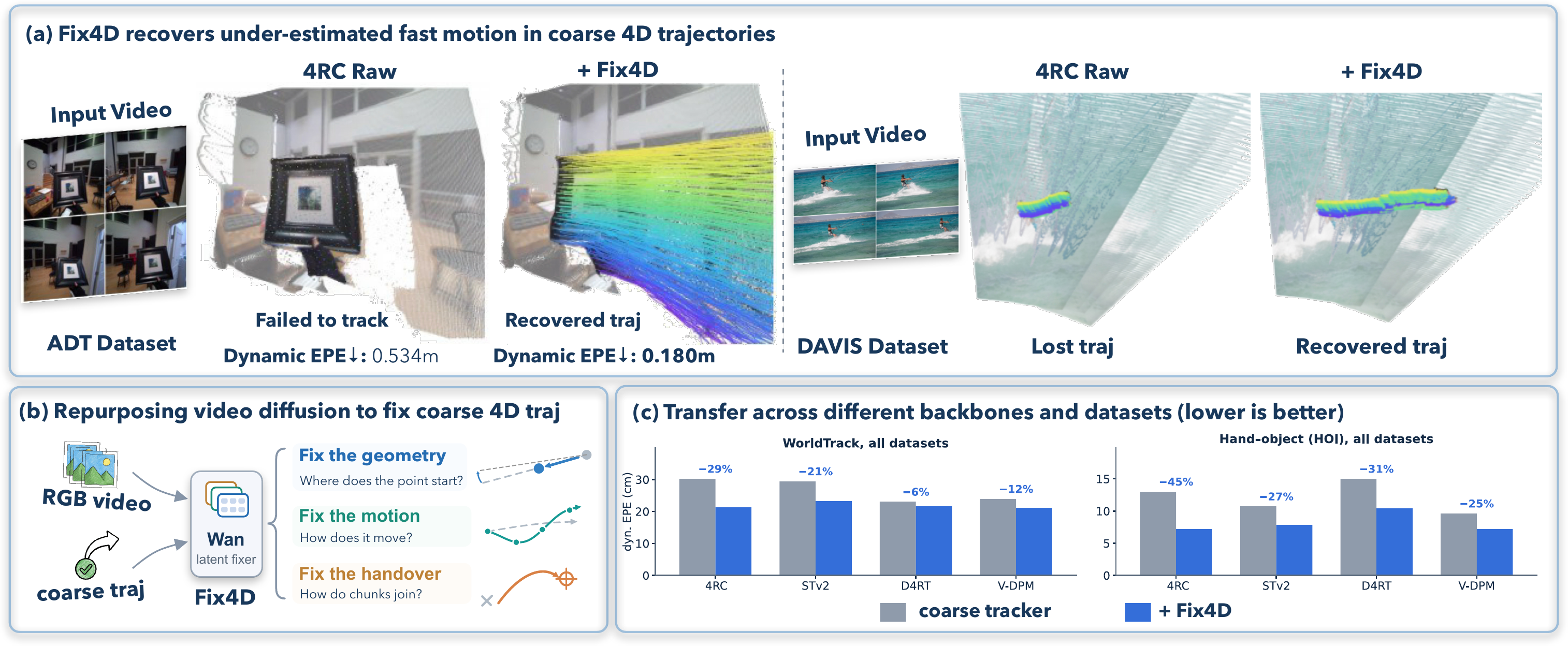}
\captionof{figure}{\textbf{Fix4D repurposes video generative
priors to correct coarse 4D trajectories.}
(a) On ADT, Fix4D recovers motion underestimated
by 4RC, largely reducing dynamic-point error; DAVIS further illustrates refinement
on in-the-wild videos.
(b) Given RGB video and coarse trajectories, Fix4D
corrects initial geometry and temporal displacement,
while adapted handover refinement corrects their composition
across chunks.
(c) Trained exclusively on 4RC proposals~\cite{luo20264rc}, the same
refiner improves dynamic-point tracking across four
backbones~\cite{zhang2026efficiently,xiao2025spatialtrackerv2,luo20264rc,sucar2026v} on WorldTrack and HOI benchmarks without retraining.}
\label{fig:teaser}
  \end{center}
}]

\begin{abstract}
Feed-forward 4D reconstruction methods recover geometry
and motion from monocular video, but their trajectories
can start at inaccurate 3D positions and underestimate
displacement, especially during rapid articulation and
object manipulation.
We introduce \textbf{Fix4D}, which repurposes video
generative priors to refine existing 4D trajectories
rather than predict them anew.
Coarse trajectories supply point identities, first-frame
geometry, and subsequent displacements, providing a
structured reference for correcting placement and
motion errors.
Conditioned on these trajectories and the observed RGB
video, Fix4D jointly predicts corrections to both
components in the native latent space of a frozen
video autoencoder.
Source-indexed representations preserve point identities,
while projection-guided feature fusion connects them
to spatiotemporal RGB evidence.
The decoded corrections update the coarse trajectories
in a single deterministic forward pass.
We further adapt the refiner for association and overlap
correction at chunk handovers in long videos.
Trained only on 4RC proposals, a single Fix4D model
reduces average dynamic-point error across four tracking
backbones without retraining, including on entirely
held-out datasets, and achieves state-of-the-art 4D
tracking performance on both general-scene and
hand--object interaction benchmarks.
Even from static trajectory inputs, Fix4D recovers
missing motion, improving tracking beyond geometry
correction alone, while informative coarse motion
yields further gains.
\end{abstract}    

\section{Introduction}
\label{sec:intro}

Understanding how the 3D world evolves over time requires
both reconstructing scene structure and following points
as they move, articulate, and interact.
This is particularly challenging in hand--object interaction
(HOI), where rapid articulation, object manipulation,
and frequent occlusion demand accurate spatial and
temporal reasoning.

Existing approaches differ in how they connect
point identity, geometry, and motion.
The first line builds tracking on dense per-frame geometry
from off-the-shelf 3D reconstruction
models~\cite{wang2026vggtomega,lin2025depth}.
Methods such as TAPIP3D~\cite{zhang2026tapip3d},
TrackEverything~\cite{jain2026trackeverything}, and
TrackCraft3R~\cite{nam2026trackcraft3r} infer which
3D locations across these geometric snapshots
correspond to the same physical point.
Such associations can be ambiguous under articulation
and occlusion, while using per-frame pre-estimated geometry limits holistic joint
refinement of geometry and motion. The second line jointly models static and dynamic
scene content through persistent 3D point trajectories,
as in D4RT~\cite{zhang2026efficiently},
4RC~\cite{luo20264rc}, and V-DPM~\cite{sucar2026v}.
Each trajectory follows the same source point across
time: its 3D position in the first frame defines
the reference geometry, and its position in each
subsequent frame can be expressed as that initial
position plus a 3D displacement.
These predictions can still place points at incorrect
3D positions and underestimate their displacement,
particularly during rapid hand articulation and
object manipulation.

In parallel, pretrained generative models have been
repurposed for depth and 3D motion
estimation~\cite{ke2024repurposing,zhang2026dvd,zhu2026motioncrafter}.
This raises a complementary question:
\emph{can video generative priors improve existing 4D
estimates rather than predicting trajectories anew?}
Persistent point indexing lets us organize corrections
as video-like fields, providing a natural interface
to video generative priors.
However, correcting the first-frame position only
relocates the trajectory as a whole; recovering
missing motion requires correcting subsequent
displacements.
Building on this representation, we
\emph{factorize the correction, not the estimation},
jointly predicting both updates from coarse
trajectories and video evidence.

We introduce \textbf{Fix4D}, a video-native framework
that refines coarse 4D trajectories from a frozen
tracker using the observed RGB video.
Geometry and displacement corrections are represented
as source-indexed fields in the native latent space
of a frozen Wan video autoencoder~\cite{wan2025wan}.
Two video-pretrained transformer towers combine
the coarse proposals with RGB evidence:
a correction tower predicts updates conditioned
on the trajectories, while projection-guided fusion
connects its source-indexed tokens to spatiotemporal
features from the RGB tower.
The decoded corrections update first-frame geometry
and subsequent displacements in a single deterministic
forward pass.
Operating on explicit trajectories rather than
tracker-specific features provides a shared refinement
interface across 4D estimators.
For long videos, we further adapt the refiner to
chunk handovers, improving moving-point association
and overlap placement when composing trajectory segments.

Trained exclusively on 4RC~\cite{luo20264rc} proposals,
Fix4D transfers to
SpatialTrackerV2~\cite{xiao2025spatialtrackerv2},
D4RT~\cite{zhang2026efficiently}, and
V-DPM~\cite{sucar2026v} without retraining.
It achieves state-of-the-art 4D tracking performance
on WorldTrack and HOI and improves average dynamic-point
accuracy across all four evaluated backbones,
including on held-out datasets.
Ablations show that motion correction improves tracking
beyond geometry refinement alone, even from static inputs.
On the evaluated HOI subset, static-input refinement
approaches refinement of full 4RC trajectories,
while informative coarse motion provides further gains.
Our contributions are summarized as:
\begin{itemize}
    \item \textbf{Joint, video-native 4D correction.}
    We jointly predict corrections to the first-frame
    geometry and subsequent displacements of coarse
    trajectories in a shared video-native latent space,
    using source-indexed representations and
    projection-guided video feature fusion.

    \item \textbf{Transferable trajectory refinement.}
    A single model trained on 4RC proposals improves
    average dynamic-point accuracy across diverse tracking
    backbones without retraining, including on entirely
    held-out datasets.

    \item \textbf{Extension to long-video tracking.}
    We adapt the refiner to chunk handovers, combining
    moving-point association and overlap correction
    with gated merging of successive trajectory segments.
\end{itemize}
\section{Related Work}
\label{sec:related_work}
\textbf{4D reconstruction and point tracking.}
D4RT~\cite{zhang2026efficiently} independently decodes
3D point queries across pixels, times, and camera
reference frames from a video encoder initialized
with VideoMAEv2~\cite{wang2023videomae}.
4RC~\cite{luo20264rc} adds a time-conditioned
displacement decoder to a geometry backbone initialized
from Depth Anything 3, while V-DPM~\cite{sucar2026v}
extends VGGT with a decoder that maps points observed
in different frames to their positions at a common
target time.
SpatialTrackerV2~\cite{xiao2025spatialtrackerv2}
combines learned video depth with coupled 2D/3D track
updates and differentiable bundle adjustment of camera
poses.
TAPIP3D~\cite{zhang2026tapip3d} lifts image features
into 3D using supplied depth and updates trajectories
through 3D neighborhood attention.
Fix4D uses RGB and explicit geometry and motion
proposals as a shared refinement interface across
these estimation architectures.\\
\textbf{Diffusion priors for geometric perception.}
Marigold~\cite{ke2024repurposing} fine-tunes Stable
Diffusion for RGB-conditioned depth denoising, while
DVD~\cite{zhang2026dvd} adapts video diffusion for
single-pass regression from RGB to depth latents.
DiTracker~\cite{son2025probing} fuses matching costs
from LoRA-adapted video-DiT query--key features and
ResNet features for 2D tracking.
For dense 3D tracking,
TrackCraft3R~\cite{nam2026trackcraft3r} combines RGB
and given per-frame pointmaps with reference-anchored track
latents in a video diffusion transformer.
MotionCrafter~\cite{zhu2026motioncrafter} learns
dedicated geometry and motion VAEs to reconstruct
pointmaps and scene flow.
ArtiFixer~\cite{de2026artifixer} uses
diffusion-generated views to refine 3D reconstructions.
Fix4D predicts geometry and motion corrections
to supplied trajectories through the native latent
space of a frozen video autoencoder.\\
\textbf{Long-video tracking and trajectory fusion.}
For causal 2D tracking,
Track-On~\cite{aydemir2025trackon} maintains
appearance and query-history memories, while
TAPNext~\cite{zholus2025tapnext} combines attention
over image and point tokens with temporal
state-space recurrence.
In 3D, Point4D~\cite{jeon2026point4d} re-queries
predicted positions across $\mathrm{Sim}(3)$-aligned
overlapping windows while retaining appearance
descriptors.
TrackEverything~\cite{jain2026trackeverything}
maintains persistent world-space tracks, separately
refines endpoints and dynamic trajectories, and
merges co-located endpoints at window boundaries.
At handovers, Fix4D learns to recover unlinked
moving-point correspondences and correct overlapping
trajectory segments before gated merging.
\section{Method}
\label{sec:method}
\begin{figure*}[htbp]
\centering
\includegraphics[width=0.98\linewidth]{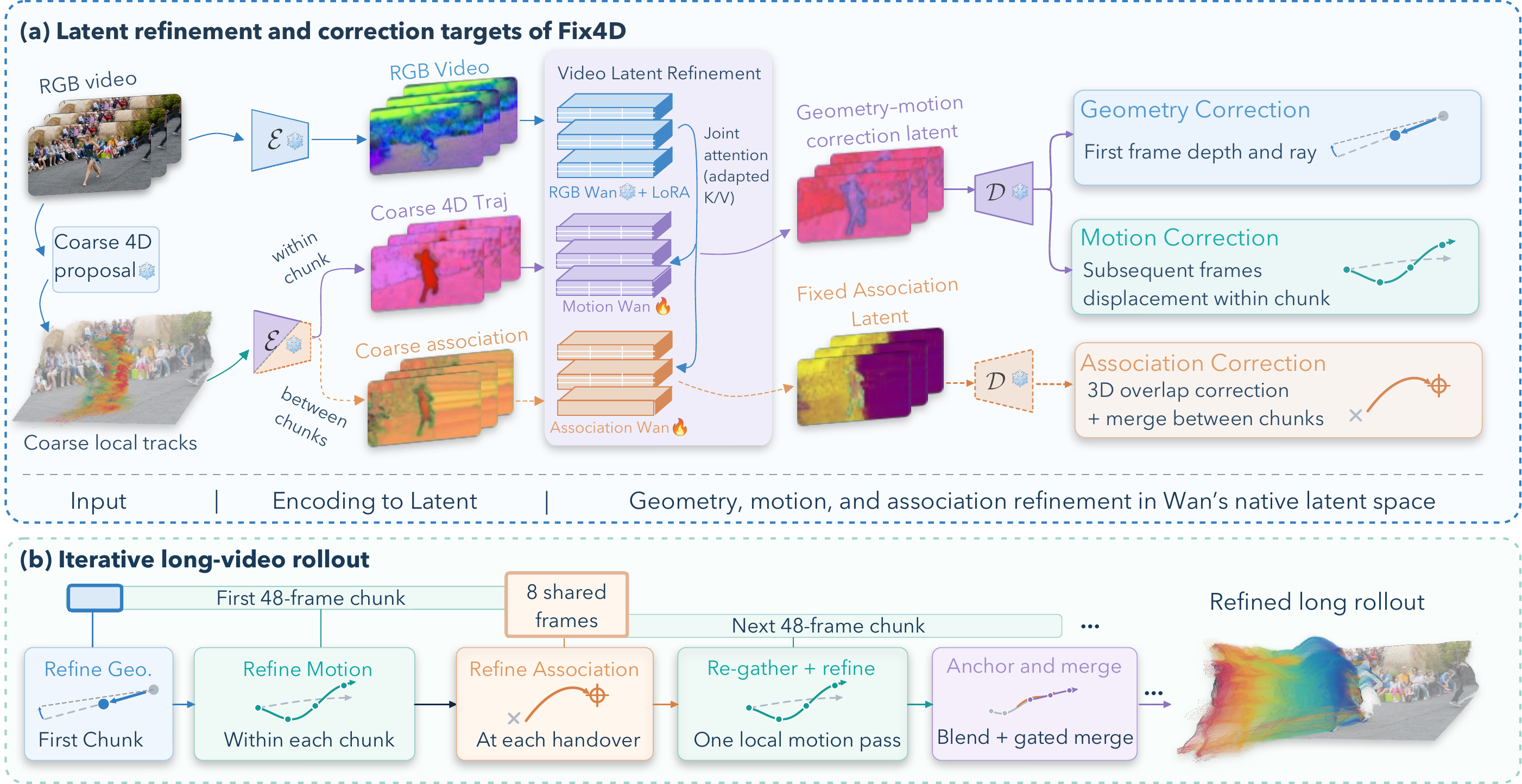}
\caption{\textbf{Fix4D architecture.}
(a) The local refiner jointly predicts corrections
to first-frame geometry and subsequent 3D displacements
from coarse trajectories and RGB video.
The decoded corrections update the supplied trajectories.
(b) Long-video rollout applies geometry correction
only in the first chunk and motion correction in
every chunk.
An adapted association tower refines moving-point
associations and overlap placement at chunk handovers.}
\label{fig:method_two_stage}
\end{figure*}

Starting from an RGB video, we employ a frozen
4D reconstruction model, such as 4RC~\cite{luo20264rc},
to estimate dense 3D point trajectories, camera
parameters, and validity information. Fix4D uses the same RGB observations to refine these
coarse predictions by learning corrections to initial
geometry and temporal displacement in the native
latent space of a pretrained video model.
Our framework first performs local geometry and motion
refinement within each video window, then extends
video-conditioned correction to learned fusion of
overlapping tracklets for long-video tracking
(Fig.~\ref{fig:method_two_stage}).

\subsection{Factorized correction of 4D trajectories}
\label{sec:motion_problem}
We represent the coarse 4D tracks as persistent trajectories
$\mathbf{P}_i(t)$, $t=0,\dots,T-1$, where point $i$ is indexed by its
source-image location and all positions are expressed in the first
camera frame. Each trajectory is the sum of its \emph{first-frame
geometry} $\mathbf{p}^c_i=\mathbf{P}_i(0)$ and its \emph{subsequent
displacement} $\mathbf{m}^c_i(t)=\mathbf{P}_i(t)-\mathbf{P}_i(0)$, with
$\mathbf{m}^c_i(0)=\mathbf{0}$. Fix4D predicts one correction for each:
\begin{equation}
\hat{\mathbf{X}}_i(t)=
\underbrace{\mathbf{p}^c_i+\Delta\hat{\mathbf{p}}_i}_{\text{first-frame geometry}}
+\underbrace{\mathbf{m}^c_i(t)+\hat{\mathbf{r}}_i(t)}_{\text{subsequent displacement}}.
\label{eq:factorized_correction}
\end{equation}
In training and inference, we set
$\widehat{\mathbf r}_i(0)=\mathbf 0$, so only the
geometry correction acts at $t=0$.
The geometry correction $\Delta\widehat{\mathbf p}_i$
is a single offset per point: it sets where the
trajectory starts and shifts all its frames together.
The displacement correction
$\widehat{\mathbf r}_i(t)$, $t\geq 1$, adjusts the
point's displacement relative to this corrected start.
The coarse terms come from the frozen tracker,
and Fix4D predicts both corrections jointly.
\subsection{Video-native latent representation}
\label{sec:correction_latent}

We organize coarse 3D displacements as
three-channel spatiotemporal fields indexed by the
source-image grid. Let $\mathbf u_i$ denote the
source-grid location of tracked point $i$.
At each time $t$, we represent its displacement as
\begin{equation}
\mathcal M_t^c(\mathbf u_i)=\mathbf m_i^c(t).
\label{eq:motion_field}
\end{equation}
The same representation is used for predicted motion
corrections. Each spatial location retains the identity
of the same tracked point throughout a video window,
rather than following its changing image projection.
This preserves consistent point indexing while
representing temporal 3D motion as a video-like signal,
even when a tracked point moves outside the image.
We encode these fields using the frozen Wan video
autoencoder~\cite{wan2025wan}.
To accommodate the numerical range of 3D displacements,
we apply reversible signed square-root companding
with a per-window scale derived from the coarse motion.
Let $\mathcal E_a$ and $\mathcal D_a$ denote the
corresponding encoding and decoding operations,
including this normalization. We obtain
\begin{equation}
z^c=\mathcal E_a(\mathcal M^c),
\qquad
\widehat{\mathcal R}
=\mathcal D_a(\widehat z_{\mathrm{mot}}),
\label{eq:motion_codec}
\end{equation}
where $z^c$ encodes the displacement representation
of the coarse 4D trajectories,
$\widehat z_{\mathrm{mot}}$ is the motion-correction
component of the joint geometry--motion output
$\widehat z$, and $\widehat{\mathcal R}$ is the
decoded 3D displacement-correction field. We represent initial geometry separately from temporal
displacement. For each source point, we parameterize
its coarse 3D position using depth $d_i^c$ and
normalized image-plane ray $\mathbf q_i^c$.
The network predicts a log-depth correction
$\widehat\gamma_i$ and ray offset
$\widehat{\boldsymbol\rho}_i$, yielding the
refined initial position
\begin{equation}
\widehat{\mathbf p}_i^0
=
d_i^c e^{\widehat\gamma_i}
[\mathbf q_i^c+\widehat{\boldsymbol\rho}_i;1].
\label{eq:initial_geometry}
\end{equation}
This gives $\Delta\hat{\mathbf{p}}_i=\hat{\mathbf{p}}^0_i-\mathbf{p}^c_i$
in Eq.~\eqref{eq:factorized_correction}, and the displacement correction is read from the decoded
field, $\hat{\mathbf{r}}_i(t)=\hat{\mathcal{R}}_t(\mathbf{u}_i)$ for
$t\geq 1$.
The geometry corrections form a separate three-channel
image on the source grid, encoded using the same
frozen autoencoder. Its latent slice is appended to
the motion latents, allowing initial geometry
and temporal displacement corrections to be predicted
jointly while retaining their distinct geometric
interpretations.

\subsection{RGB-conditioned two-tower refinement}
\label{sec:motion_conditioning}

To exploit visual evidence when refining coarse trajectories,
we employ two Wan video denoising transformer
towers~\cite{wan2025wan}, initialized from a
video-depth-adapted pretrained checkpoint~\cite{zhang2026dvd}.
The RGB tower processes the observed RGB sequence and
extracts spatiotemporal visual features. The correction
tower operates on source-indexed correction tokens and
predicts geometry and motion residuals in Wan's native
latent space. It starts from a zero-initialized correction
latent, conditioned on the encoded coarse motion,
source-frame appearance, projected image coordinates,
relative depth, camera rays, and validity information. The coarse trajectories provide the geometric reference, while the RGB tower supplies visual evidence for correcting their inaccuracies.\\
\textbf{Projection-guided RGB feature fusion.}
The two towers use different spatial references.
RGB tokens retain their image-grid positions, whereas
correction tokens maintain fixed source-grid identities.
To connect these representations, we project the coarse
3D trajectories into each observed frame. Since the video
autoencoder compresses spatial and temporal dimensions,
we aggregate valid projected locations through
validity-weighted spatial and temporal pooling to obtain
rotary positions at the latent-token resolution.
Tokens without valid projections retain their source-grid
positions.

At every transformer block, correction queries attend
jointly to their own features and those from the RGB tower.
Let $h_\ell$ and $f_\ell$ contain the correction
and RGB tokens as rows at block $\ell$, respectively.
Suppressing normalization and positional rotations
for clarity, the attention operation is
\begin{equation}
\begin{aligned}
K_\ell &=
[h_\ell W_k^\top;\,
 f_\ell(W_k+B_kA_k)^\top],\\
V_\ell &=
[h_\ell W_v^\top;\,
 f_\ell(W_v+B_vA_v)^\top],\\
o_\ell &= \operatorname{softmax}\!\left(
\frac{(h_\ell W_q^\top)K_\ell^\top}{\sqrt{d_h}}
\right)V_\ell.
\end{aligned}
\label{eq:motion_attention}
\end{equation}
RGB features extend the attention keys and values,
allowing visual information to flow into the correction
stream without updating the RGB tokens through this
fusion operation. Here, $W_q,W_k,W_v$ are query, key, and value
projection matrices, $A_k,B_k,A_v,B_v$ are trainable
low-rank factors, $d_h$ is the attention-head
dimension, and $o_\ell$ is the correction-token
attention output.\\
\textbf{Latent prediction and adaptation.}
The two-tower prediction process can be summarized as
\begin{equation}
\begin{aligned}
F_{\rm RGB} = g_\phi(I),\quad 
\widehat z = f_\theta
\left(\mathbf 0;\,z^c,F_{\rm RGB},c\right),
\end{aligned}
\label{eq:latent_prediction}
\end{equation}
where $I$ is the RGB video, $g_\phi$ and $f_\theta$
are the RGB and correction towers, and $c$ collects
additional geometric conditioning. The geometry and motion components of $\widehat z$
are decoded separately and applied to the coarse
trajectories through
Eq.~\eqref{eq:factorized_correction}.

We retain pretrained visual representations while adapting them to trajectory correction.
The video autoencoder and the RGB tower's base weights
remain frozen, while the correction tower is fully trained.
We introduce rank-128 LoRA adapters~\cite{hu2021lora}
in the RGB tower's self-attention projections and
rank-64 key/value adapters for the RGB features accessed
by the correction tower, as in
Eq.~\eqref{eq:motion_attention}.
The adapters are initialized to introduce zero residual
updates, preserving the pretrained initialization at
the start of training.

Unlike iterative video generation, Fix4D predicts
the correction latent deterministically in a single
forward pass at a fixed timestep. The decoded corrections
yield the refined geometry and motion in ~\cref{eq:factorized_correction}.
Further architectural details are provided in the
supplementary material.

\subsection{Handover refinement for long videos}
\label{sec:tracklet_fusion}

Long videos are processed in 48-frame chunks with
eight-frame overlaps.
The baseline composition links carried points to
incoming tracklets through their projected positions,
then anchors the tracklets at the last shared frame
and blends them over the overlap.
Two errors can arise.
First, occlusion, drift, or out-of-view projections
can produce incorrect or missing continuations,
leaving points without reliable cross-chunk associations.
Second, anchoring incoming tracklets to the carried
trajectory can preserve existing position errors
and propagate them into subsequent frames.

We address both errors with an \emph{association tower},
a separately adapted copy of the trained local refiner
that operates only at handovers.
It takes the raw incoming proposal $\mathbf P$,
carried trajectories $\mathbf H$, and RGB observations
over the eight shared frames and the first new frame.
For unlinked moving points, it selects a continuation
among nearby candidate tracks, or abstains.
For linked moving points, it predicts a corrected
3D overlap $\mathbf C$ and a per-point gate $g_i\in[0,1]$.
Accepted associations update the incoming tracks
before the chunk's single local motion-refinement
pass; initial geometry is refined only in the first chunk.
Let $\mathbf B$ denote the anchored and blended
composition of the locally refined chunk with
$\mathbf H$.
For a chunk starting at global frame $\tau$,
$\mathcal O=\{\tau,\ldots,\tau+7\}$ indexes the
shared frames.
For points receiving overlap corrections, we apply
\begin{equation}
\mathbf Y_i(t)=\mathbf B_i(t)
+g_i\bigl(\mathbf C_i(t)-\mathbf B_i(t)\bigr),
\quad t\in\mathcal O.
\label{eq:learned_merge}
\end{equation}
The final overlap offset is propagated to subsequent
frames of the chunk, preserving their frame-to-frame
displacements.
Points predicted static retain the baseline
composition ($g_i=0$), and frames before the overlap
remain unchanged.
Detailed association, composition, and training
procedures are provided in the supplementary material.

\subsection{Learning objectives}
\label{sec:motion_learning}
For dense trajectories, we encode geometrically defined
correction fields with the frozen video autoencoder
to obtain latent targets.
Let $\mathbf X_i^\star(t)$ denote the ground-truth
3D position of point $i$ at time $t$.
The motion targets are
$\mathbf r_i^\star(t)=
[\mathbf X_i^\star(t)-\mathbf X_i^\star(0)]
-\mathbf m_i^c(t)$ for $t\geq 1$.
The geometry targets are
$\gamma_i^\star=\log(d_i^\star/d_i^c)$ and
$\boldsymbol\rho_i^\star=\mathbf q_i^\star-\mathbf q_i^c$,
where $d_i^\star$ and $\mathbf q_i^\star$ are the
depth and normalized ray of $\mathbf X_i^\star(0)$,
recovering its initial position via
Eq.~\eqref{eq:initial_geometry}.

Sparse annotations cannot provide complete latent targets,
so we supervise decoded 3D predictions at annotated
point-time pairs. Geometric supervision also complements
latent regression on dense examples. We use
\begin{equation}
\begin{aligned}
\mathcal L_{\rm latent}
&= \|\widehat z-z^\star\|_{\Omega_z}^2,\\
\mathcal L_{\rm sparse}
&= \frac{1}{|\mathcal S|}
\sum_{(i,t)\in\mathcal S}
\ell\bigl(\widehat{\mathbf X}_i(t),
          \mathbf X_i^\star(t)\bigr),
\end{aligned}
\label{eq:local_objective}
\end{equation}
where $z^\star$ denotes the encoded correction target,
$\Omega_z$ is the valid latent support, and
$\mathcal S$ contains annotated point-time pairs.
The function $\ell$ measures geometric reconstruction
error in 3D.
The handover model is subsequently adapted using
supervision on assembled trajectories.
Additional loss terms
are detailed in the supplementary material.

\section{Experiments}
\label{sec:experiment}
\subsection{Training and Evaluation Setup}
\begin{figure*}[htbp]
\centering
\includegraphics[width=0.98\linewidth]{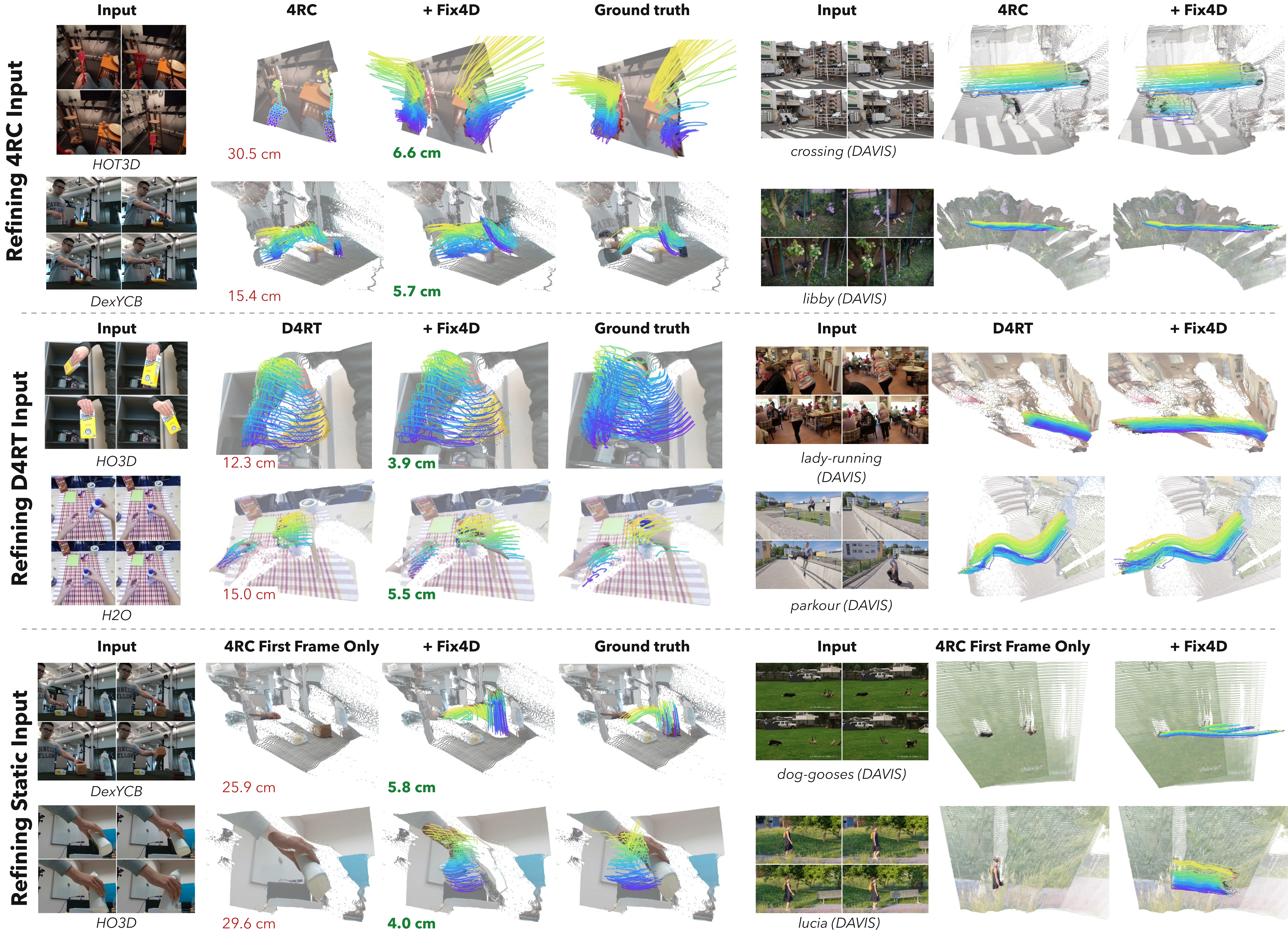}
\caption{\textbf{Qualitative 4D tracking refinement.}
Fix4D refines trajectories from 4RC~\cite{luo20264rc} (top)
and D4RT~\cite{zhang2026efficiently} (middle) without retraining, and recovers
motion from static 4RC~\cite{luo20264rc} inputs (bottom).
Left: HOI examples with ground truth and
dynamic-point EPE (cm) of the displayed tracks.
Right: in-the-wild DAVIS~\cite{perazzi2016benchmark} examples.}
\label{fig:quali}
\end{figure*}
\textbf{Benchmarks.} We evaluate Fix4D on general-scene tracking and
hand--object interaction (HOI).
\textbf{WorldTrack.}
We follow WorldTrack~\cite{feng2025st4rtrack} and the
D4RT~\cite{zhang2026efficiently} evaluation protocol
on 172 videos from ADT~\cite{pan2023aria},
Dynamic Replica~\cite{karaev2023dynamicstereo},
PointOdyssey~\cite{zheng2023pointodyssey}, and
Panoptic Studio~\cite{joo2015panoptic}.
For HOI, we construct a benchmark of 401 clips from six datasets:
HO3D~\cite{hampali2020honnotate},
H2O~\cite{kwon2021h2o},
DexYCB~\cite{chao2021dexycb},
ARCTIC~\cite{fan2023arctic},
HOT3D~\cite{banerjee2025hot3d},
TACO~\cite{liu2024taco}.
We derive 3D trajectories from annotated hand and object
poses, including background points where available.
HO3D~\cite{hampali2020honnotate},
H2O~\cite{kwon2021h2o}, and ARCTIC~\cite{fan2023arctic} are held-out datasets,
while DexYCB~\cite{chao2021dexycb},
HOT3D~\cite{banerjee2025hot3d}, and
TACO~\cite{liu2024taco} use held-out evaluation
sequences from training-covered datasets. We evaluate 48-frame windows on both WorldTrack and HOI benchmarks.
For long-video tracking, we use the first 300 frames
of 20 Dynamic Replica, 20 PointOdyssey, and 16 HO3D
videos (Table~\ref{tab:long}).\\
\textbf{Training details.}
We use frozen 4RC~\cite{luo20264rc} to generate
coarse 4D trajectories on synthetic and real-world
sequences, and train Fix4D under 3D supervision.
Our training data include
Syn4D~\cite{jiang2026syn4d},
Dynamic Replica~\cite{karaev2023dynamicstereo},
PointOdyssey~\cite{zheng2023pointodyssey},
HOT3D~\cite{banerjee2025hot3d},
OakInk2~\cite{zhan2024oakink2}, TACO~\cite{liu2024taco} and
DexYCB~\cite{chao2021dexycb}.
We employ temporal and motion augmentations to expose
the model to diverse dynamics and varying levels of
error in coarse trajectory proposals.
Training the local refiner takes approximately
25 hours on 64 NVIDIA GH200 GPUs.
The association tower is subsequently adapted from the trained local refiner
for about 11 hours on 16 GH200 GPUs.\\
\textbf{Evaluation metrics.}
We follow the 3D trajectory evaluation protocols
of~\cite{jeon2026point4d,feng2025st4rtrack,karhade2026any4d}.
Predictions are aligned to ground truth using one
median-norm scale per sequence, without rotation or
translation fitting.
We report end-point error (EPE) and average percentage
within distance thresholds (APD) on all points and dynamic
points only. Scores are averaged over sequences within
each dataset and then equally across datasets.
Metric definitions and dynamic-point criteria are
detailed in the supplementary material.\\
\textbf{Comparisons and transfer protocol.}
We compare Fix4D against sparse point trackers
and dense 4D reconstruction methods under matched
query and evaluation protocols.
To assess transferability, the same refiner trained
on 4RC proposals is applied to
SpatialTrackerV2~\cite{xiao2025spatialtrackerv2},
D4RT~\cite{zhang2026efficiently}, and
V-DPM~\cite{sucar2026v} without retraining.
For D4RT, we use a publicly available community
reimplementation, as the official code is unavailable.
Provider outputs are converted into compatible
source-indexed conditioning fields, and predicted
corrections are applied to each provider's own
trajectories.

Additional implementation and evaluation details are provided in the supplementary material.
\subsection{Results}
\begin{table*}[t]
\centering
\setlength{\tabcolsep}{3pt}
\caption{\textbf{General 3D tracking on WorldTrack~\cite{feng2025st4rtrack}.}
APD (\%, $\uparrow$) and EPE (m, $\downarrow$) on all
and dynamic points over 48 frames, with one median scale
per sequence. Avg is the mean across four datasets.
A single Fix4D checkpoint refines
4RC~\cite{luo20264rc}, STv2~\cite{xiao2025spatialtrackerv2},
D4RT~\cite{zhang2026efficiently}, and V-DPM~\cite{sucar2026v}.
\first{Best}, \second{second}, and \third{third}
results are highlighted.}
\label{tab:general}
\resizebox{0.9\textwidth}{!}{%
\begin{tabular}{lcccccccccc}
\toprule
 & \multicolumn{2}{c}{ADT} & \multicolumn{2}{c}{DR} & \multicolumn{2}{c}{PO} & \multicolumn{2}{c}{PStudio} & \multicolumn{2}{c}{Avg} \\
\cmidrule(lr){2-3} \cmidrule(lr){4-5} \cmidrule(lr){6-7} \cmidrule(lr){8-9} \cmidrule(lr){10-11}
 & APD$\uparrow$ & EPE$\downarrow$ & APD$\uparrow$ & EPE$\downarrow$ & APD$\uparrow$ & EPE$\downarrow$ & APD$\uparrow$ & EPE$\downarrow$ & APD$\uparrow$ & EPE$\downarrow$ \\
Method & All / Dyn. & All / Dyn. & All / Dyn. & All / Dyn. & All / Dyn. & All / Dyn. & All / Dyn. & All / Dyn. & All / Dyn. & All / Dyn. \\
\midrule
\multicolumn{11}{l}{\emph{Sparse tracking}} \\
STv2~\cite{xiao2025spatialtrackerv2} & \first{91.7} / 74.4 & \second{0.097} / 0.271 & 69.4 / 67.7 & 0.302 / 0.352 & 85.6 / 74.7 & 0.170 / 0.322 & 74.4 / 74.4 & 0.231 / 0.232 & 80.3 / 72.8 & 0.200 / 0.294 \\
TAPIP3D~\cite{zhang2026tapip3d} & \third{87.8} / \third{75.9} & 0.140 / \third{0.262} & 67.8 / 68.8 & 0.308 / 0.333 & 76.3 / 59.4 & 0.220 / 0.410 & 75.6 / 75.9 & 0.204 / 0.202 & 76.9 / 70.0 & 0.218 / 0.302 \\
\midrule
\multicolumn{11}{l}{\emph{Dense tracking}} \\
MonST3R~\cite{zhang2025monst3r} & 76.1 / 54.0 & 0.251 / 0.448 & 58.0 / 53.3 & 0.441 / 0.564 & 67.4 / 53.4 & 0.317 / 0.509 & 55.3 / 55.2 & 0.413 / 0.414 & 64.2 / 54.0 & 0.355 / 0.484 \\
St4RTrack~\cite{feng2025st4rtrack} & 74.2 / 64.2 & 0.282 / 0.361 & 74.3 / 73.3 & 0.260 / 0.233 & 75.1 / 70.5 & 0.224 / 0.249 & 71.7 / 72.0 & 0.239 / 0.237 & 73.8 / 70.0 & 0.251 / 0.270 \\
Any4D~\cite{karhade2026any4d} & 59.0 / 55.6 & 0.399 / 0.435 & \third{79.6} / 71.5 & \third{0.209} / 0.301 & 76.6 / 70.3 & 0.208 / 0.285 & 65.2 / 65.1 & 0.284 / 0.285 & 70.1 / 65.6 & 0.275 / 0.327 \\
TraceAnything~\cite{liu2026trace} & 77.7 / 62.1 & 0.230 / 0.353 & 60.7 / 61.4 & 0.571 / 0.455 & 53.5 / 44.7 & 0.672 / 0.630 & 72.8 / 73.0 & 0.261 / 0.259 & 66.2 / 60.3 & 0.433 / 0.424 \\
TrackCraft3R~\cite{nam2026trackcraft3r} & \second{90.1} / \first{78.0} & \third{0.113} / \second{0.227} & 71.0 / 59.7 & 0.308 / 0.456 & 85.0 / 71.3 & 0.158 / 0.296 & 62.3 / 62.3 & 0.317 / 0.317 & 77.1 / 67.8 & 0.224 / 0.324 \\
Point4D~\cite{jeon2026point4d} & 84.5 / 72.1 & 0.262 / 0.296 & 77.8 / 73.5 & 0.234 / 0.304 & 83.9 / 75.2 & 0.178 / 0.253 & 69.7 / 70.3 & 0.264 / 0.258 & 79.0 / 72.8 & 0.235 / 0.278 \\
V-DPM~\cite{sucar2026v} & 86.1 / 61.1 & 0.162 / 0.422 & 75.6 / 74.9 & 0.242 / 0.233 & \second{90.2} / \second{87.0} & \second{0.111} / \second{0.129} & \third{79.6} / \third{79.6} & \third{0.171} / \third{0.171} & \second{82.8} / \third{75.7} & \third{0.172} / 0.239 \\
D4RT~\cite{zhang2026efficiently} & 73.8 / 56.9 & 0.259 / 0.378 & 76.7 / \second{79.7} & 0.244 / 0.210 & 74.7 / 80.5 & 0.226 / 0.175 & \first{80.9} / \first{81.8} & \second{0.167} / \first{0.157} & 76.5 / 74.7 & 0.224 / 0.230 \\
4RC~\cite{luo20264rc} & 85.5 / 54.2 & 0.170 / 0.474 & \second{82.3} / 76.9 & \second{0.184} / 0.265 & \third{90.0} / 81.4 & 0.117 / 0.204 & 68.1 / 68.5 & 0.267 / 0.264 & 81.5 / 70.3 & 0.184 / 0.302 \\
\midrule
\multicolumn{11}{l}{\emph{Fix4D (ours)}} \\
4RC~\cite{luo20264rc} + Fix4D & \third{87.8} / 68.0 & 0.140 / 0.302 & \first{88.8} / \first{82.5} & \first{0.131} / \first{0.170} & \first{92.8} / \third{86.4} & \first{0.085} / \third{0.131} & 69.6 / 69.7 & 0.251 / 0.250 & \first{84.8} / \second{76.6} & \first{0.152} / \second{0.213} \\
STv2~\cite{xiao2025spatialtrackerv2} + Fix4D & \first{91.7} / \second{76.0} & \first{0.094} / \first{0.225} & 73.0 / 71.6 & 0.254 / 0.268 & 87.2 / 78.9 & 0.141 / 0.220 & 75.3 / 75.4 & 0.213 / 0.213 & 81.8 / 75.5 & {0.176} / 0.231 \\
D4RT~\cite{zhang2026efficiently} + Fix4D & 73.5 / 62.1 & 0.259 / 0.336 & 76.7 / \third{78.7} & 0.237 / \second{0.197} & 74.1 / 80.2 & 0.228 / 0.172 & \second{80.8} / \second{81.5} & \first{0.164} / \first{0.157} & 76.3 / 75.6 & 0.222 / \third{0.216} \\
V-DPM~\cite{sucar2026v} + Fix4D & 86.0 / 64.6 & 0.157 / 0.340 & 76.1 / 75.8 & 0.229 / \third{0.209} & 89.1 / \first{87.5} & \third{0.113} / \first{0.119} & \third{79.6} / 79.5 & \third{0.171} / \second{0.170} & \third{82.7} / \first{76.8} & \second{0.167} / \first{0.210} \\
\bottomrule
\end{tabular}
}
\end{table*}
\begin{table*}[t]
\centering
\setlength{\tabcolsep}{3pt}
\caption{\textbf{Hand--object interaction tracking on
HO3D~\cite{hampali2020honnotate},
H2O~\cite{kwon2021h2o},
DexYCB~\cite{chao2021dexycb},
ARCTIC~\cite{fan2023arctic},
HOT3D~\cite{banerjee2025hot3d}, and
TACO~\cite{liu2024taco}.}
APD (\%, $\uparrow$) and EPE (m, $\downarrow$) over
48 frames, reported for all/dynamic points with
one median scale per sequence.
Avg is the six-dataset mean.
The same Fix4D checkpoint refines
4RC~\cite{luo20264rc},
STv2~\cite{xiao2025spatialtrackerv2},
D4RT~\cite{zhang2026efficiently}, and
V-DPM~\cite{sucar2026v}.
\first{Best}, \second{second}, and \third{third}
are highlighted.}
\label{tab:hoi}
\resizebox{0.89\textwidth}{!}{%
\begin{tabular}{lcccccccccccccc}
\toprule
 & \multicolumn{2}{c}{HO3D} & \multicolumn{2}{c}{H2O} & \multicolumn{2}{c}{DexYCB} & \multicolumn{2}{c}{ARCTIC} & \multicolumn{2}{c}{HOT3D} & \multicolumn{2}{c}{TACO} & \multicolumn{2}{c}{Avg} \\
\cmidrule(lr){2-3} \cmidrule(lr){4-5} \cmidrule(lr){6-7} \cmidrule(lr){8-9} \cmidrule(lr){10-11} \cmidrule(lr){12-13} \cmidrule(lr){14-15}
 & APD$\uparrow$ & EPE$\downarrow$ & APD$\uparrow$ & EPE$\downarrow$ & APD$\uparrow$ & EPE$\downarrow$ & APD$\uparrow$ & EPE$\downarrow$ & APD$\uparrow$ & EPE$\downarrow$ & APD$\uparrow$ & EPE$\downarrow$ & APD$\uparrow$ & EPE$\downarrow$ \\
Method & All / Dyn. & All / Dyn. & All / Dyn. & All / Dyn. & All / Dyn. & All / Dyn. & All / Dyn. & All / Dyn. & All / Dyn. & All / Dyn. & All / Dyn. & All / Dyn. & All / Dyn. & All / Dyn. \\
\midrule
\multicolumn{15}{l}{\emph{Sparse tracking}} \\
STv2~\cite{xiao2025spatialtrackerv2} & \third{84.0} / 84.5 & \third{0.162} / 0.154 & \first{95.2} / 95.2 & \first{0.072} / 0.068 & 92.8 / 90.2 & 0.142 / 0.178 & 96.8 / 96.6 & 0.054 / 0.058 & 81.4 / 87.4 & 0.199 / 0.129 & 97.7 / 97.0 & 0.050 / 0.055 & \third{91.3} / 91.8 & 0.113 / 0.107 \\
TAPIP3D~\cite{zhang2026tapip3d} & 82.1 / 78.0 & 0.188 / 0.220 & 89.9 / \third{96.7} & 0.102 / 0.064 & \third{94.8} / 92.6 & 0.106 / 0.139 & 97.7 / 97.7 & 0.044 / \second{0.045} & 64.4 / 72.5 & 0.349 / 0.225 & 98.4 / 98.1 & \third{0.045} / 0.048 & 87.9 / 89.2 & 0.139 / 0.123 \\
\midrule
\multicolumn{15}{l}{\emph{Dense tracking}} \\
MonST3R~\cite{zhang2025monst3r} & 76.5 / 77.5 & 0.221 / 0.220 & 71.0 / 75.8 & 0.237 / 0.193 & 89.5 / 84.4 & 0.120 / 0.163 & 91.8 / 89.5 & 0.090 / 0.106 & 73.2 / 74.9 & 0.263 / 0.210 & 89.7 / 87.6 & 0.097 / 0.105 & 82.0 / 81.6 & 0.171 / 0.166 \\
St4RTrack~\cite{feng2025st4rtrack} & 76.1 / 76.8 & 0.227 / 0.213 & 77.9 / 89.3 & 0.188 / 0.100 & 91.0 / 87.2 & 0.109 / 0.142 & 95.3 / 94.2 & 0.063 / 0.068 & 69.3 / 78.0 & 0.295 / 0.174 & 90.3 / 90.1 & 0.096 / 0.097 & 83.3 / 85.9 & 0.163 / 0.132 \\
Any4D~\cite{karhade2026any4d} & 83.4 / 80.5 & 0.171 / 0.194 & 82.8 / 86.5 & 0.142 / 0.114 & 87.6 / 83.6 & 0.281 / 0.369 & 96.1 / 95.0 & 0.056 / 0.062 & 64.6 / 74.7 & 0.325 / 0.202 & 98.1 / 97.8 & 0.049 / 0.052 & 85.4 / 86.4 & 0.170 / 0.165 \\
TraceAnything~\cite{liu2026trace} & 70.2 / 72.6 & 0.364 / 0.272 & 82.6 / 92.1 & 0.195 / 0.086 & 86.6 / 84.2 & 0.130 / 0.150 & 94.4 / 93.9 & 0.077 / 0.083 & 68.0 / 74.2 & 0.363 / 0.222 & 91.7 / 89.6 & 0.088 / 0.099 & 82.2 / 84.4 & 0.203 / 0.152 \\
TrackCraft3R~\cite{nam2026trackcraft3r} & 79.2 / 75.1 & 0.200 / 0.222 & \third{93.2} / 96.5 & \second{0.084} / \third{0.063} & 89.8 / 86.4 & 0.179 / 0.228 & \first{98.7} / \first{98.7} & \second{0.039} / \first{0.041} & 79.7 / 88.9 & 0.211 / 0.116 & 98.4 / 98.0 & \third{0.045} / 0.050 & 89.8 / 90.6 & 0.127 / 0.120 \\
Point4D~\cite{jeon2026point4d} & 80.1 / 81.9 & 0.200 / 0.187 & 88.4 / 95.2 & 0.122 / 0.070 & 85.1 / 85.2 & 0.250 / 0.280 & \third{98.3} / 97.8 & \third{0.041} / \second{0.045} & 80.2 / 85.1 & 0.213 / 0.140 & 97.1 / 96.1 & 0.057 / 0.064 & 88.2 / 90.2 & 0.147 / 0.131 \\
V-DPM~\cite{sucar2026v} & 83.7 / 83.0 & 0.164 / 0.171 & 84.2 / 96.4 & 0.134 / 0.067 & 94.5 / 91.9 & 0.091 / 0.120 & 98.2 / 98.0 & 0.047 / 0.049 & 81.7 / 89.5 & 0.197 / 0.113 & 98.4 / 98.2 & 0.056 / 0.059 & 90.1 / 92.8 & 0.115 / \third{0.096} \\
D4RT~\cite{zhang2026efficiently} & 81.5 / 79.9 & 0.198 / 0.186 & 90.6 / 83.3 & 0.099 / 0.127 & 84.9 / 83.2 & 0.253 / 0.280 & 96.2 / 95.2 & 0.057 / 0.061 & 75.3 / 76.7 & 0.239 / 0.178 & 96.8 / 96.7 & 0.067 / 0.069 & 87.5 / 85.8 & 0.152 / 0.150 \\
4RC~\cite{luo20264rc} & 79.4 / 72.9 & 0.203 / 0.265 & 92.4 / 96.5 & 0.095 / \third{0.063} & 93.3 / 90.6 & \third{0.090} / \third{0.114} & 97.4 / 96.0 & 0.046 / 0.055 & 75.6 / 73.8 & 0.246 / 0.229 & \third{98.6} / 98.3 & 0.050 / 0.053 & 89.4 / 88.0 & 0.122 / 0.130 \\
\midrule
\multicolumn{15}{l}{\emph{Fix4D (ours)}} \\
4RC~\cite{luo20264rc} + Fix4D & 81.5 / 80.5 & 0.183 / 0.180 & 92.7 / \first{97.6} & \third{0.092} / \first{0.055} & \first{97.0} / \first{96.0} & \first{0.051} / \first{0.058} & \second{98.6} / \third{98.1} & \first{0.037} / \first{0.041} & \first{87.2} / \first{95.9} & \first{0.144} / \first{0.066} & \first{99.7} / \first{99.7} & \first{0.033} / \first{0.035} & \first{92.8} / \third{94.6} & \first{0.090} / \first{0.072} \\
STv2~\cite{xiao2025spatialtrackerv2} + Fix4D & \second{85.8} / \first{89.2} & \second{0.153} / \first{0.114} & \second{93.3} / \second{96.8} & \second{0.084} / \second{0.060} & 94.7 / \third{93.1} & 0.095 / 0.119 & 98.2 / 98.0 & 0.045 / 0.047 & \second{85.8} / \third{92.7} & \second{0.157} / \second{0.086} & \second{99.1} / \third{98.8} & \second{0.037} / \third{0.041} & \first{92.8} / \second{94.8} & \second{0.095} / \second{0.078} \\
D4RT~\cite{zhang2026efficiently} + Fix4D & 83.7 / \third{85.7} & 0.167 / \third{0.135} & 92.1 / 91.4 & 0.094 / 0.086 & 89.2 / 88.6 & 0.142 / 0.163 & 97.6 / 97.5 & 0.051 / 0.053 & 81.3 / 89.1 & 0.191 / \third{0.105} & 95.3 / 94.6 & 0.075 / 0.078 & 89.9 / 91.2 & 0.120 / 0.104 \\
V-DPM~\cite{sucar2026v} + Fix4D & \first{86.4} / \second{88.7} & \first{0.146} / \second{0.122} & 86.2 / \first{97.6} & 0.127 / \second{0.060} & \second{96.1} / \second{94.8} & \second{0.068} / \second{0.081} & \first{98.7} / \second{98.6} & 0.044 / \third{0.046} & \third{85.2} / \second{93.3} & \third{0.165} / \second{0.086} & \first{99.7} / \second{99.6} & \second{0.037} / \second{0.039} & \second{92.1} / \first{95.5} & \third{0.098} / \first{0.072} \\
\bottomrule
\end{tabular}
}
\end{table*}
\begin{table}[t]
\centering
\caption{\textbf{Cross-tracker transfer of Fix4D.}
Dynamic point EPE (cm, $\downarrow$) over 48 frames.
\emph{In domain} denotes held-out sequences from training-covered
datasets while \emph{out-of-domain} denotes entirely
held-out datasets.}
\label{tab:providers}
\small
\setlength{\tabcolsep}{4pt}
\resizebox{0.8\linewidth}{!}{%
\begin{tabular}{lcccc}
\toprule
 & \multicolumn{2}{c}{HOI} & \multicolumn{2}{c}{WorldTrack} \\
\cmidrule(lr){2-3}\cmidrule(lr){4-5}
Provider & Out of domain & In domain & Out of domain & In domain \\
\midrule
4RC~\cite{luo20264rc} (training provider) & 12.8 & 13.2 & 36.9 & 23.5 \\
\quad \textbf{+ Fix4D} & \textbf{9.2} & \textbf{5.3} & \textbf{27.6} & \textbf{15.1} \\
\qquad $\Delta$ & $-$28\% & $-$60\% & $-$25\% & $-$36\% \\
\midrule
STv2~\cite{xiao2025spatialtrackerv2} & 9.3 & 12.1 & 25.2 & 33.7 \\
\quad \textbf{+ Fix4D} & \textbf{7.4} & \textbf{8.2} & \textbf{21.9} & \textbf{24.4} \\
\qquad $\Delta$ & $-$21\% & $-$32\% & $-$13\% & $-$28\% \\
\midrule
D4RT~\cite{zhang2026efficiently} & 12.5 & 17.6 & 26.8 & 19.2 \\
\quad \textbf{+ Fix4D} & \textbf{9.1} & \textbf{11.6} & \textbf{24.7} & \textbf{18.5} \\
\qquad $\Delta$ & $-$27\% & $-$34\% & $-$8\% & $-$4\% \\
\midrule
V-DPM~\cite{sucar2026v} & 9.6 & 9.7 & 29.7 & 18.1 \\
\quad \textbf{+ Fix4D} & \textbf{7.6} & \textbf{6.9} & \textbf{25.5} & \textbf{16.4} \\
\qquad $\Delta$ & $-$21\% & $-$29\% & $-$14\% & $-$9\% \\
\bottomrule
\end{tabular}%
}
\end{table}

\begin{table*}[t]
\centering
\setlength{\tabcolsep}{3pt}
\caption{\textbf{Long-video 3D tracking.}
Dynamic-point APD (\%, $\uparrow$) and EPE (m, $\downarrow$)
over the first 300 frames and their final 48 on
DR~\cite{karaev2023dynamicstereo},
PO~\cite{zheng2023pointodyssey}, and
HO3D~\cite{hampali2020honnotate}.
Our variants process these sequences in 48-frame chunks,
using local refinement alone or combined with handover refinement.
\first{Best}, \second{second}, and \third{third} are highlighted.}
\label{tab:long}
\resizebox{0.8\textwidth}{!}{%
\begin{tabular}{lcccccccc}
\toprule
 & \multicolumn{2}{c}{DR} & \multicolumn{2}{c}{PO} & \multicolumn{2}{c}{HO3D} & \multicolumn{2}{c}{Avg} \\
\cmidrule(lr){2-3} \cmidrule(lr){4-5} \cmidrule(lr){6-7} \cmidrule(lr){8-9}
 & APD$\uparrow$ & EPE$\downarrow$ & APD$\uparrow$ & EPE$\downarrow$ & APD$\uparrow$ & EPE$\downarrow$ & APD$\uparrow$ & EPE$\downarrow$ \\
Method & Full / Last 48 & Full / Last 48 & Full / Last 48 & Full / Last 48 & Full / Last 48 & Full / Last 48 & Full / Last 48 & Full / Last 48 \\
\midrule
\multicolumn{9}{l}{\emph{Sparse tracking}} \\
TAPIP3D~\cite{zhang2026tapip3d} & 70.9 / 66.6 & 0.308 / 0.353 & 47.6 / 39.3 & 1.329 / 1.916 & 77.2 / \third{77.1} & 0.248 / 0.252 & 65.2 / 61.0 & 0.628 / 0.841 \\
\midrule
\multicolumn{9}{l}{\emph{Dense tracking}} \\
MonST3R~\cite{zhang2025monst3r} & 51.1 / 48.6 & 0.552 / 0.595 & 19.6 / 17.3 & 1.987 / 2.382 & 67.6 / 66.5 & 0.327 / 0.325 & 46.1 / 44.1 & 0.956 / 1.100 \\
St4RTrack~\cite{feng2025st4rtrack} & \third{74.9} / \first{72.7} & \first{0.215} / \first{0.233} & 31.5 / 28.8 & 1.198 / 1.648 & 70.7 / 71.1 & 0.256 / 0.247 & 59.0 / 57.5 & 0.556 / 0.710 \\
Any4D~\cite{karhade2026any4d} & 65.7 / 61.6 & 0.350 / 0.392 & 41.9 / 35.6 & 1.001 / 1.543 & \first{81.7} / \first{78.5} & \third{0.198} / \second{0.210} & 63.1 / 58.6 & 0.516 / 0.715 \\
TrackCraft3R~\cite{nam2026trackcraft3r} & 61.0 / 60.9 & 0.400 / 0.410 & 35.6 / 32.3 & 1.500 / 1.955 & 76.2 / 74.9 & 0.224 / \third{0.234} & 57.6 / 56.0 & 0.708 / 0.866 \\
Point4D~\cite{jeon2026point4d} & 71.4 / 68.0 & 0.304 / 0.334 & 44.1 / 37.4 & 1.049 / 1.550 & 74.2 / 70.7 & 0.264 / 0.293 & 63.2 / 58.7 & 0.539 / 0.726 \\
V-DPM~\cite{sucar2026v} & 69.5 / 65.1 & 0.273 / 0.314 & \third{52.2} / \third{43.2} & \third{0.769} / \third{1.307} & 75.9 / 71.8 & 0.228 / 0.264 & 65.9 / 60.1 & 0.423 / 0.628 \\
4RC~\cite{luo20264rc} & 69.3 / 64.0 & 0.312 / 0.359 & 47.6 / 40.3 & 1.021 / 1.549 & 66.6 / 62.6 & 0.354 / 0.384 & 61.2 / 55.7 & 0.562 / 0.764 \\
\midrule
\multicolumn{9}{l}{\emph{Ours}} \\
4RC~\cite{luo20264rc} + Fix4D (local only) & \second{75.2} / \third{68.9} & \third{0.227} / \third{0.278} & 50.0 / 42.0 & 0.804 / 1.349 & 74.1 / 71.0 & 0.250 / 0.273 & 66.4 / 60.6 & 0.427 / 0.633 \\
4RC~\cite{luo20264rc} + Fix4D (w.\ handover refinement) & \first{76.2} / \second{71.0} & \second{0.217} / \second{0.258} & 50.3 / 42.2 & 0.799 / 1.344 & 75.0 / 73.8 & 0.244 / 0.254 & \third{67.2} / \second{62.3} & \third{0.420} / \third{0.619} \\
V-DPM~\cite{sucar2026v} + Fix4D (local only) & 70.2 / 65.3 & 0.252 / 0.296 & \second{53.1} / \second{44.5} & \second{0.754} / \second{1.293} & \third{80.9} / 76.5 & \second{0.176} / \second{0.210} & \second{68.1} / \third{62.1} & \second{0.394} / \second{0.599} \\
V-DPM~\cite{sucar2026v} + Fix4D (w.\ handover refinement) & 70.4 / 65.9 & 0.250 / 0.292 & \first{53.4} / \first{44.6} & \first{0.746} / \first{1.279} & \second{81.3} / \second{77.5} & \first{0.173} / \first{0.201} & \first{68.4} / \first{62.7} & \first{0.390} / \first{0.591} \\
\bottomrule
\end{tabular}
}
\end{table*}

Fix4D reduces average dynamic-point EPE and improves average dynamic-point APD for
all four backbones on WorldTrack and HOI
(Tables~\ref{tab:general} and~\ref{tab:hoi}).
Trained only on 4RC~\cite{luo20264rc} proposals,
it transfers to
SpatialTrackerV2~\cite{xiao2025spatialtrackerv2},
D4RT~\cite{zhang2026efficiently}, and
V-DPM~\cite{sucar2026v} without retraining,
achieving overall state-of-the-art 4D tracking
performance on both benchmarks.
\Cref{fig:quali} illustrates cross-backbone refinement
and motion recovery from static inputs on HOI, with
substantial reductions in dynamic-point error.
Examples on DAVIS~\cite{perazzi2016benchmark}
further demonstrate qualitative recovery of missing
or underestimated motion across diverse
in-the-wild videos.

Beyond transfer across tracking backbones, Fix4D
also generalizes to unseen datasets
(Table~\ref{tab:providers}).
Across all four input providers, refinement consistently
reduces dynamic-point EPE on both in-domain held-out
sequences and entirely held-out datasets.
On unseen datasets, relative error reductions range
from $21$--$28\%$ on HOI and $8$--$25\%$ on WorldTrack.
These results demonstrate that the learned video-native
corrections generalize beyond both the tracking backbone
and the scene distributions.

Long-video tracking also requires accurate trajectory composition across window handovers.
On 300-frame sequences, local refinement improves
tracking for both 4RC and V-DPM, while handover
refinement further reduces dynamic-point EPE
(Table~\ref{tab:long}).
By correcting trajectory association and alignment
across chunk boundaries, handover refinement
reduces accumulated errors, yielding larger
improvements over the final 48 frames.

\subsection{Ablation Studies}
\label{sec:ablation}
\begin{table}[t]
\centering
\setlength{\tabcolsep}{3pt}
\caption{\textbf{Geometry and motion correction.}
Average APD (\%, $\uparrow$) and EPE (m, $\downarrow$)
on all and dynamic points of the 48-frame benchmarks.
The same checkpoint is evaluated with no correction,
geometry correction alone, and joint geometry--motion
correction.}
\label{tab:ablation_components}
\resizebox{0.9\columnwidth}{!}{%
\begin{tabular}{cc cccc cccc}
\toprule
\multicolumn{2}{c}{Correction} & \multicolumn{4}{c}{General (avg.)} & \multicolumn{4}{c}{HOI (avg.)} \\
\cmidrule(lr){1-2}\cmidrule(lr){3-6}\cmidrule(lr){7-10}
& & \multicolumn{2}{c}{APD$\uparrow$} & \multicolumn{2}{c}{EPE$\downarrow$} & \multicolumn{2}{c}{APD$\uparrow$} & \multicolumn{2}{c}{EPE$\downarrow$} \\
Geometry & Motion & All & Dyn. & All & Dyn. & All & Dyn. & All & Dyn. \\
\midrule
-- & -- & 81.5 & 70.3 & 0.184 & 0.302 & 89.4 & 88.0 & 0.122 & 0.130 \\
\checkmark & -- & 84.0 & 73.6 & 0.159 & 0.247 & 92.4 & 93.3 & 0.093 & 0.082 \\
\checkmark & \checkmark & \textbf{84.8} & \textbf{76.6} & \textbf{0.152} & \textbf{0.213} & \textbf{92.8} & \textbf{94.6} & \textbf{0.090} & \textbf{0.072} \\
\bottomrule
\end{tabular}}
\end{table}

\begin{table}[t]
\centering
\caption{\textbf{RGB tower adaptation}
on WorldTrack and HOI, reported as All / Dynamic.
We compare raw 4RC with Fix4D using a frozen
RGB Wan tower or LoRA adaptation.}
\label{tab:ablation_rgb}
\small
\setlength{\tabcolsep}{4pt}
\renewcommand{\arraystretch}{1.1}
\resizebox{0.9\linewidth}{!}{%
\begin{tabular}{lcccc}
\toprule
& \multicolumn{2}{c}{WorldTrack}
& \multicolumn{2}{c}{HOI} \\
\cmidrule(lr){2-3}\cmidrule(lr){4-5}
Method & APD$\uparrow$ & EPE$\downarrow$
       & APD$\uparrow$ & EPE$\downarrow$ \\
\midrule
4RC (raw)
& 81.5 / 70.3 & 0.184 / 0.302
& 89.4 / 88.0 & 0.122 / 0.130 \\
4RC + Fix4D (frozen RGB Wan)
& 84.1 / 76.4 & 0.159 / 0.217
& 90.9 / 92.0 & 0.107 / 0.095 \\
\textbf{4RC + Fix4D (LoRA RGB Wan)}
& \textbf{84.8 / 76.6} & \textbf{0.152 / 0.213}
& \textbf{92.8 / 94.6} & \textbf{0.090 / 0.072} \\
\bottomrule
\end{tabular}%
}
\end{table}

\begin{table}[t]
\centering
\caption{\textbf{Motion recovery from static proposals.}
Average APD (\%, $\uparrow$) and EPE (m, $\downarrow$)
on general-scene and HOI benchmarks.
All / Dyn.\ denotes all and dynamic points, excluding ADT~\cite{pan2023aria}, Hot3D~\cite{banerjee2025hot3d} sets, where 4RC~\cite{luo20264rc} heavily underpredicts motion.}
\label{tab:ablation_static}
\small
\setlength{\tabcolsep}{3pt}
\renewcommand{\arraystretch}{1.12}
\resizebox{0.9\columnwidth}{!}{%
\begin{tabular}{lcccc}
\toprule
& \multicolumn{2}{c}{General}
& \multicolumn{2}{c}{HOI} \\
\cmidrule(lr){2-3}\cmidrule(lr){4-5}
Method
& APD$\uparrow$ & EPE$\downarrow$
& APD$\uparrow$ & EPE$\downarrow$ \\
\midrule
\multicolumn{5}{l}{\textit{Static input:
4RC frame 0 repeated (no motion)}} \\
4RC (raw)
& 73.2 / 64.4 & 0.256 / 0.349
& 90.5 / 86.7 & 0.107 / 0.135 \\
4RC + Fix4D (geometry only)
& 75.8 / 67.1 & 0.228 / 0.295
& 91.9 / 89.2 & 0.093 / 0.106 \\
\textbf{4RC + Fix4D}
& \textbf{82.0 / 76.6} & \textbf{0.170 / 0.207}
& \textbf{93.7 / 93.9} & \textbf{0.082 / 0.078} \\
\midrule
\multicolumn{5}{l}{\textit{Full input:
original 48-frame 4RC trajectories}} \\
4RC (raw)
& 80.2 / 75.6 & 0.189 / 0.245
& 92.2 / 90.9 & 0.097 / 0.110 \\
\textbf{4RC + Fix4D}
& \textbf{83.7 / 79.5} & \textbf{0.156 / 0.184}
& \textbf{93.9 / 94.4} & \textbf{0.079 / 0.074} \\
\bottomrule
\end{tabular}%
}
\end{table}

\textbf{Geometry and motion correction.}
We compare unrefined trajectories, geometry-only
correction, and joint geometry--motion correction,
with both correction variants using the same trained
model (Table~\ref{tab:ablation_components}).
On WorldTrack, geometry correction reduces dynamic-point
EPE from $0.302$ to $0.247$\,m, and adding motion
correction further reduces it to $0.213$\,m.
On HOI, geometry correction reduces EPE from
$0.130$ to $0.082$\,m, while joint correction reaches
$0.072$\,m---a $45\%$ reduction relative to the
unrefined trajectories.
These results show that correcting initial placement
improves tracking, while motion correction provides
additional gains beyond geometry refinement alone.\\
\textbf{Motion recovery from static proposals.}
We investigate whether Fix4D can recover missing motion
without an initial motion hypothesis, and whether
informative coarse trajectories further improve refinement.
We construct static inputs by repeating the initial
3D positions across all frames (Table~\ref{tab:ablation_static}).
Since geometry correction alone cannot recover temporal
displacement, we compare it against joint correction.
On the evaluated subsets, enabling motion correction
reduces dynamic-point EPE from $0.295$ to $0.207$\,m
on general scenes and from $0.106$ to $0.078$\,m
on HOI. Qualitative examples in Fig.~3 further
illustrate motion recovered from static inputs.
On the evaluated HOI subset, static-input refinement
approaches refinement of full 4RC trajectories
($0.078$ vs.\ $0.074$\,m).
Nevertheless, informative coarse motion yields further
gains on both benchmarks, reaching $0.184$\,m on
the WorldTrack subsets and $0.074$\,m on HOI subsets.
These results highlight the ability of video-pretrained
representations to recover missing motion while
benefiting from coarse trajectory estimates.\\
\textbf{Frozen vs.\ adapted video features.}
We investigate whether pretrained video representations
can support 4D refinement without adapting the RGB
backbone (Table~\ref{tab:ablation_rgb}).
Remarkably, keeping the RGB Wan tower frozen already
yields substantial improvements, reducing dynamic-point
EPE from $0.302$ to $0.217$\,m on WorldTrack and
from $0.130$ to $0.095$\,m on HOI.
This demonstrates that pretrained video features
provide effective visual guidance for learning
geometry and motion corrections, even without
updating the RGB tower.
Our final LoRA-adapted configuration further improves
performance, particularly on HOI, reaching $0.072$\,m.
\subsection{Runtime Analysis}
Fix4D refines a 48-frame window at $528 \times 304$
resolution in 9.0\,s on an RTX~4090, compared with
9.9\,s for the 4RC~\cite{luo20264rc} inference;
Fix4D peaks at 14.4\,GiB of GPU memory.

\section{Conclusion}
We presented Fix4D, which repurposes video generative
priors to refine existing 4D trajectories rather than
reconstruct them anew.
Using coarse trajectories as structured references,
it jointly corrects first-frame geometry and subsequent
displacements in a video-native latent space.
Trained only on 4RC proposals, a single model reduces
average dynamic-point error across four tracking
backbones without retraining, including on held-out
datasets, and achieves state-of-the-art 4D tracking
performance on WorldTrack and HOI.
Fix4D recovers missing motion even from static inputs,
while informative coarse motion yields further gains.
Handover refinement reduces accumulated
error in long videos.
By operating on trajectories rather than tracker-specific
features, Fix4D provides a modular refinement approach
that can complement future advances in 4D estimation.\\
\textbf{Limitations.}
Fix4D relies on camera estimates and point
correspondences provided by the initial tracker,
and may not fully recover from severe reconstruction
errors.

{
    \small
    \bibliographystyle{ieeenat_fullname}
    \bibliography{main}
}
\clearpage
\renewcommand{\thesection}{\Alph{section}}
\setcounter{section}{0}
\setcounter{table}{0}
\renewcommand{\thetable}{S\arabic{table}}
\setcounter{figure}{0}
\renewcommand{\thefigure}{S\arabic{figure}}
\setcounter{equation}{0}
\renewcommand{\theequation}{S\arabic{equation}}
\renewcommand{\theHsection}{supp.\Alph{section}}
\renewcommand{\theHtable}{supp.\arabic{table}}
\renewcommand{\theHfigure}{supp.\arabic{figure}}
\renewcommand{\theHequation}{supp.\arabic{equation}}
\twocolumn[{%
\centering
\Large
\textbf{\thetitle}\\
\vspace{0.5em}Supplementary Material \\
\vspace{1.0em}
}]

This supplement provides additional qualitative results
(\cref{supp:qual}), an early-training ablation of video
priors (\cref{supp:init}), implementation details
(\cref{supp:method,supp:handover}), evaluation protocols
(\cref{supp:eval}), and further results and limitations
(\cref{supp:results,supp:limits}).
The accompanying video presents animated qualitative results.
Equations, tables, and figures without the prefix ``S''
refer to the main paper.

\section{Additional qualitative results}
\label{supp:qual}
\begin{figure*}[p]
\centering
\includegraphics[width=0.905\linewidth]{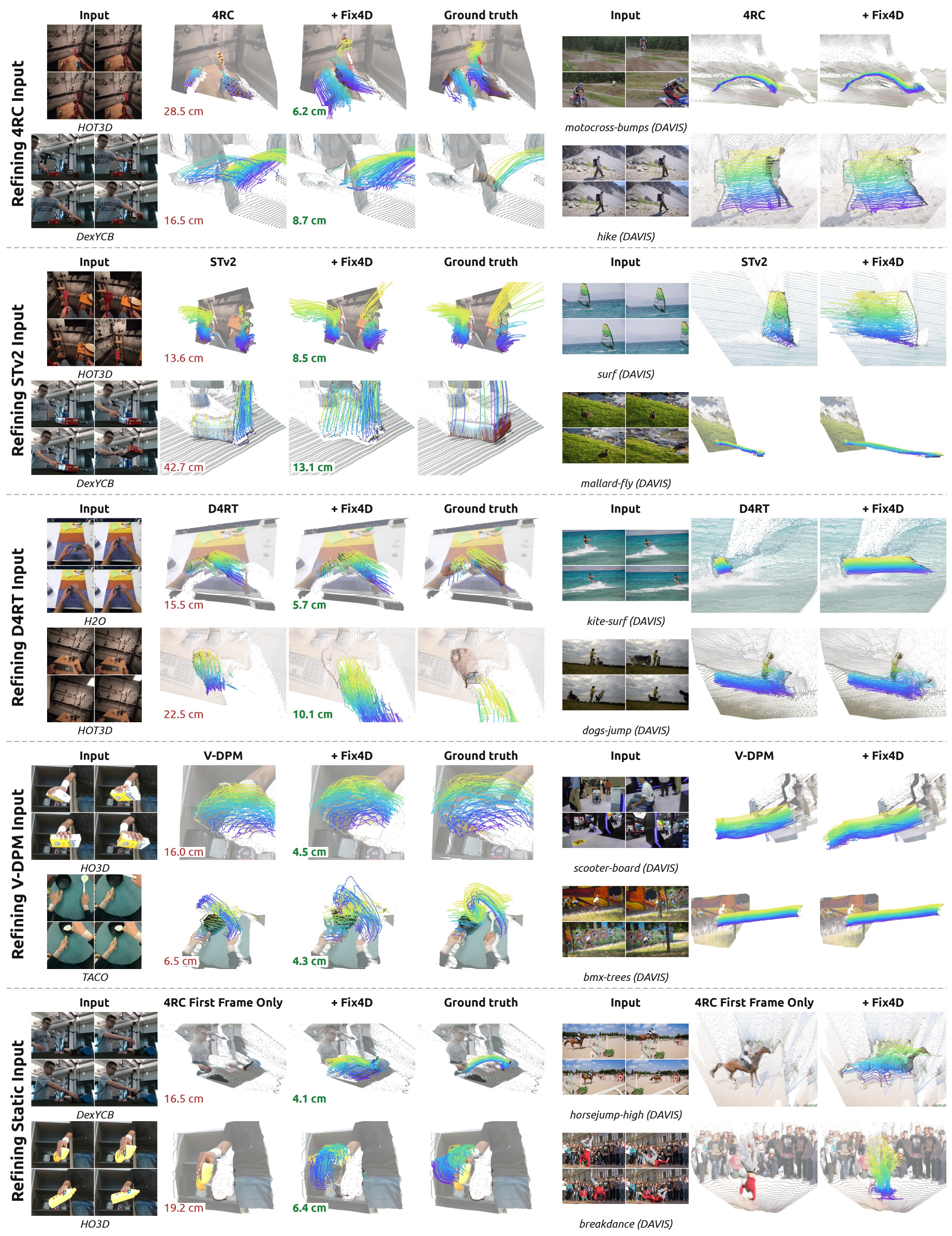}
\caption{\textbf{Additional qualitative results.}
The same Fix4D checkpoint refines trajectories from
4RC~\cite{luo20264rc},
STv2~\cite{xiao2025spatialtrackerv2},
D4RT~\cite{zhang2026efficiently}, and
V-DPM~\cite{sucar2026v} without retraining.
The bottom rows demonstrate motion recovery from static
4RC inputs: first-frame 3D positions repeated across
all 48 frames, with the RGB video unchanged.
Left: HOI examples with ground truth and dynamic-point
EPE (cm) of the displayed tracks.
Right: in-the-wild DAVIS~\cite{perazzi2016benchmark}
examples.}
\label{fig:supp_qual}
\end{figure*}

\Cref{fig:supp_qual} shows additional examples of a
single Fix4D checkpoint refining 4RC, SpatialTrackerV2,
D4RT, and V-DPM without retraining, as well as recovering
motion from static 4RC inputs.
Examples span HOI sequences with ground truth and
in-the-wild DAVIS~\cite{perazzi2016benchmark} videos.

\section{Early-training ablation of video priors}
\label{supp:init}
\begin{table*}[t]
\centering
\small
\setlength{\tabcolsep}{9pt}
\caption{\textbf{Early-training ablation of video priors.}
Short training runs of the local refiner (\cref{supp:init}).
Random initialization fails to improve the coarse trajectories
consistently, while both video-pretrained configurations
improve all reported metrics on both benchmarks.
APD (\%, $\uparrow$) and EPE (m, $\downarrow$),
reported as All / Dynamic.
\suppgain{Green} and \suppdrop{red} indicate improvement
and degradation relative to raw 4RC; best results are bold.}
\label{tab:supp_init}
\begin{tabular}{lcccc}
\toprule
& \multicolumn{2}{c}{WorldTrack}
& \multicolumn{2}{c}{HOI} \\
\cmidrule(lr){2-3}\cmidrule(lr){4-5}
Method & APD$\uparrow$ & EPE$\downarrow$
       & APD$\uparrow$ & EPE$\downarrow$ \\
\midrule
4RC (raw)
& 81.5 / 70.3
& 0.184 / 0.302
& 89.4 / 88.0
& 0.122 / 0.130 \\
\midrule
\multicolumn{5}{l}{\emph{4RC + Fix4D}} \\
\quad trained from random initialization
& \suppdrop{80.5} / \suppgain{70.5}
& \suppdrop{0.192} / \suppgain{0.299}
& \suppgain{89.8 / 88.3}
& \suppgain{0.117 / 0.125} \\
\quad frozen Wan2.2 (5B) RGB tower
& \suppgain{81.6 / 70.6}
& \suppgain{0.183 / 0.297}
& \suppgain{89.9 / 88.4}
& \suppgain{0.116 / 0.122} \\
\quad \textbf{Wan2.1 (1.3B)~\cite{zhang2026dvd} + LoRA}
& \suppgain{\textbf{81.7 / 71.1}}
& \suppgain{\textbf{0.182 / 0.288}}
& \suppgain{\textbf{91.0 / 89.6}}
& \suppgain{\textbf{0.105 / 0.111}} \\
\bottomrule
\end{tabular}
\end{table*}
\Cref{tab:supp_init} examines how video priors affect
early learning of trajectory corrections.
Each configuration is trained directly from its initialization
on 16 GH200 GPUs for 1{,}500 updates, using roughly
8\% of the GPU-hours of full training.

With both towers randomly initialized and all their
weights trained,
Fix4D fails to improve 4RC consistently, degrading
WorldTrack all-point APD from 81.5 to 80.5\%
and EPE from 0.184 to 0.192\,m.
In contrast, both video-pretrained configurations
improve all reported metrics, with the default
DVD-initialized~\cite{zhang2026dvd} configuration
performing best throughout.
The alternative uses a frozen Wan2.2 RGB
tower~\cite{wan2025wan}, accessed through learned
linear projections, while retaining the default
correction tower.
Despite keeping the RGB backbone frozen, it
outperforms random initialization across all metrics
and consistently improves upon 4RC.
Together with the RGB-tower LoRA ablation in
Tab.~\ref{tab:ablation_rgb}, these results provide further evidence
that pretrained video representations benefit
4D trajectory refinement.
\par\begingroup\dimen0=5\baselineskip \vskip 0pt plus \dimen0 \penalty-100 \vskip 0pt plus -\dimen0 \vskip\dimen0 \penalty9999 \vskip-\dimen0 \vskip 0pt\endgroup
\section{Method details}
\label{supp:method}
\subsection{Trajectories, canvas and units}
\label{supp:units}
Each point $i$ retains its source-pixel index
$\mathbf u_i$ on a $518{\times}294$ grid throughout
the window, regardless of its motion.
Trajectories are expressed in the first camera's
coordinate frame, in the provider's units.
We pad the grid to $528{\times}304$ to match the
model's spatial stride of 16; padded pixels are
marked invalid and cropped before loss computation
and output.

For training, ground-truth trajectories are expressed
in the same first-camera coordinates and divided
by the median ground-truth-to-provider depth ratio
over labeled first-frame points.
This per-clip scale aligns the targets with the
provider's units and is used only during training.

\subsection{Correction encoding}
\label{supp:codec}
\paragraph{Video autoencoder.}
We use the frozen Wan2.1 video autoencoder~\cite{wan2025wan}
(16 latent channels, temporal stride 4, spatial stride 8)
in 32-bit precision, with posterior-mode encoding and
the autoencoder's latent normalization.
A 48-frame field is padded to 49 frames by repeating
the last frame, yielding 13 latent time slices.
Decoded fields are cropped back to 48 frames, and the
displacement correction at $t=0$ is set to zero.

\paragraph{Companding.}
Each displacement component $x$ is mapped to the
autoencoder's input range using a signed square root,
\begin{equation}
\phi_a(x)=\operatorname{sgn}(x)
\min\!\bigl(\sqrt{|x|/a},\,1\bigr),\qquad
\psi_a(c)=a\,c\,|c|,
\label{eq:supp_companding}
\end{equation}
with $\psi_a$ used after decoding.
This mapping expands small displacements relative to
large ones. The range $a$ is determined per window
from a fixed high-percentile rule on the coarse
displacement magnitudes, with a lower bound proportional
to the provider's median first-frame depth.
Values beyond $\pm a$ are clipped.

\paragraph{Geometry image.}
The first-frame geometry correction is encoded as
a single image on the source grid.
Let $\bar\gamma$ denote the companded log-depth correction
(using \cref{eq:supp_companding} with a fixed range),
and $\bar{\boldsymbol\rho}=(\bar\rho_x,\bar\rho_y)$
the scaled ray correction of Eq.~(\ref{eq:initial_geometry}).
The three image channels are
\begin{equation}
\mathbf c=\bar\gamma\,\mathbf 1+
\bar\rho_x\,\mathbf b_x+\bar\rho_y\,\mathbf b_y,
\label{eq:supp_geometry_image}
\end{equation}
where $\mathbf 1=[1,1,1]^\top$ is the gray axis
and $\mathbf b_x,\mathbf b_y$ are fixed orthogonal
unit directions perpendicular to it.
As in Marigold~\cite{ke2024repurposing}, depth is
represented along the gray axis, while the two
orthogonal color directions encode ray corrections.
Decoding takes the channel mean for depth and
projections onto $\mathbf b_x,\mathbf b_y$ for rays.
Ray components are clipped to their ranges, and
the color component is scaled down when needed
to remain within the autoencoder's input range.

Encoded as a one-frame video, the geometry image
produces one latent slice, concatenated with the
13 motion slices (14 correction slices in total).
The constants of the encoding (the factors that define $a$
and the log-depth and ray ranges) are fixed across
windows and datasets, calibrated once on training
data to keep clipping and reconstruction error small.

\begin{figure*}[t]
\centering
\includegraphics[width=0.88\linewidth]{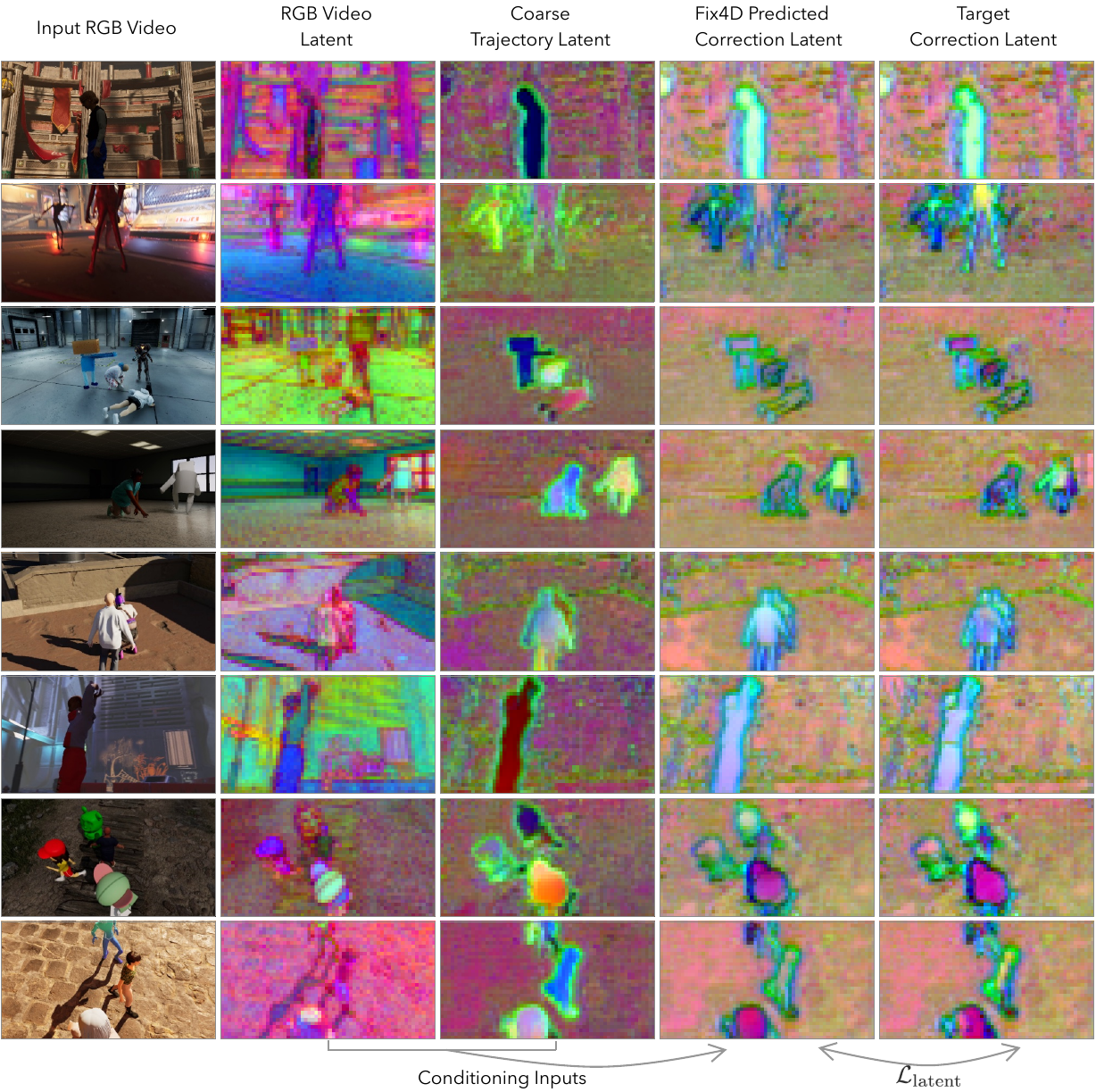}
\caption{\textbf{Correction latents on Syn4D.}
From left to right: input RGB frame, RGB video latent,
encoded coarse displacement, predicted correction latent,
and target correction latent.
The target is obtained by encoding the correction from
coarse trajectories to ground truth with the frozen
autoencoder; $\mathcal L_{\rm latent}$ (Eq.~(\ref{eq:local_objective}))
supervises this prediction.
Coarse and correction latents show one motion slice on the
source grid, indexed by first-frame pixels.
The 16-channel latents are visualized using their
first three principal components.
Each latent type uses its own PCA basis; predicted
and target corrections share the same basis and
color normalization.}
\label{fig:supp_latents}
\end{figure*}

\paragraph{Correction latents.}
Using coarse trajectories as references, we construct
explicit geometry and displacement correction fields
and encode them with the frozen video autoencoder
to obtain structured latent regression targets.
\Cref{fig:supp_latents} visualizes these targets on
Syn4D~\cite{jiang2026syn4d}: the predicted
motion-correction latents capture their main spatial
structure on the source grid.

\subsection{Architecture}
\label{supp:arch}
\paragraph{Towers.}
The local refiner comprises an RGB tower (``RGB Wan''
in \cref{fig:method_two_stage}) and a correction tower
(``Motion Wan'').
Both are Wan2.1-T2V-1.3B transformers~\cite{wan2025wan}
(30 blocks, width 1536, 12 attention heads of dimension
128, feed-forward width 8960, patch size
$1{\times}2{\times}2$), initialized from
DVD~\cite{zhang2026dvd} with an empty text prompt.
Following DVD, the RGB tower processes the video once with its base weights frozen and trainable LoRA adapters. The correction tower directly regresses correction latents in a single forward pass, conditioned on coarse proposals and RGB features. The inherited timestep inputs are fixed to 500 and 1000 for the RGB and correction towers, respectively, providing constant conditioning during training and inference. The correction-state input is fixed to zero, and the correction latent is obtained by negating the prediction head’s output.
\paragraph{Correction-tower input.}
A new patch embedding conditions the zero correction
state on source-frame appearance (the first RGB frame on
the source grid, encoded once; 16 channels),
the encoded coarse proposal (16), four validity
lanes (one per frame of the slice), and five geometric channels.
The coarse condition contains encoded displacement
on motion slices and relative log-depth on the
geometry slice.
The geometric channels comprise projected image
coordinates $(u,v)$ normalized to $[-1,1]$, relative log-depth, and
first-frame camera rays $(x/z,y/z)$, pooled with
validity weights over each latent cell and its frames.
A learned token-type vector is added to each token.
For later windows, a second zero-initialized embedding
encodes the carried-history differences
$\mathbf H-\mathbf P$ and $\mathbf H-\mathbf P(0)$
over eight shared frames, each with four validity
lanes ($16+4$ channels each).
These inputs are zero in the first window.

\paragraph{Projection-guided positions.}
The $528{\times}304$ canvas yields a
$33{\times}19$ latent patch grid, with 627 tokens
per slice, giving 8{,}778 correction tokens and
8{,}151 RGB tokens.
Correction-token rotary positions~\cite{su2024roformer}
are obtained by validity-weighted pooling of coarse
trajectory projections over each latent cell and
its frames.
Their temporal positions follow the corresponding
RGB latent slices.
Tokens without valid projections retain their
source-grid positions; geometry tokens use first-slice
projections, while RGB tokens retain their native
grid positions.

\paragraph{Fusion and adaptation.}
In each of the 30 blocks, the correction tower
reads the RGB tower's normalized and modulated
self-attention inputs through its own key/value
projections augmented with rank-64 adapters (Eq.~(\ref{eq:motion_attention})).
The resulting keys and values are concatenated with
those of the correction tokens (16{,}929 tokens total).
The correction tower retains its own cross-attention,
feed-forward layers, and output head.
The RGB tower uses rank-128 LoRA~\cite{hu2021lora}
on its self-attention query, key, value, and output
projections in blocks 1--29.
All adapters start with a zero output matrix.
Trainable parameters comprise the correction tower
and embeddings (1.420\,B), key/value adapters
(11.8\,M), and LoRA (45.6\,M).
The RGB tower's base weights (1.419\,B) and
autoencoder remain frozen.

\subsection{Training objective}
\label{supp:loss}
Equation~(8) summarizes the training objective.
With $L$ denoting the median first-frame provider
depth, we supervise the correction latents, their
decoded fields, and the composed trajectories
using the following terms:

\begin{itemize}
\item \emph{Correction latent}: masked mean squared
error against the encoded target correction latent,
with motion slices supervised on densely labeled
windows (excluding cells near unsupported pixels)
and the geometry slice supervised where depth and
ray targets are valid and representable.

\item \emph{Decoded motion}: mean Euclidean error,
normalized by $L$, between the decoded displacement
correction $\hat{\mathbf r}_i(t)$ and its target,
over labeled point-frames $t\geq1$ with valid ground
truth at both frames $0$ and $t$.

\item \emph{Decoded depth}: mean absolute error of
the log-depth correction $\hat\gamma_i$ over labeled
first-frame points.

\item \emph{Decoded ray}: mean Euclidean error against
$\boldsymbol\rho^\star_i=
\bigl(\mathbf K_{\rm gt}^{-1}[\mathbf u_i;1]\bigr)_{1:2}
-\mathbf q^c_i$
(equivalently $\mathbf q^\star_i-\mathbf q^c_i$
at labeled points).
This target requires only ground-truth intrinsics
and is available densely in each training window.

\item \emph{Composed position}: weighted Huber
loss~\cite{huber1992robust} on
$\|\hat{\mathbf X}_i(t)-\mathbf X^\star_i(t)\|/L$
over labeled point-frames, including $t=0$.
Errors are averaged separately over moving and
static points, then equally across the two groups.
The Huber threshold is set to the median normalized
error on training windows, and the term's weight
balances its gradient norm against that of the
decoded-motion term; both are calibrated once when
the term is introduced.
\end{itemize}

The correction-latent term forms $\mathcal L_{\rm latent}$
in Eq.~(\ref{eq:local_objective}), while the composed-position term is
$\mathcal L_{\rm sparse}$.
The decoded losses back-propagate through the frozen
autoencoder decoder.
The ray correction is introduced in stage~2
(\cref{supp:training}) with a factor ramped linearly
from zero to one over 200 updates; the factor is
one at inference.

\subsection{Training data and schedule}
\label{supp:training}

\paragraph{Training stages.}
Stage 1 starts from the DVD initialization (\cref{supp:arch})
by training the correction tower for 4{,}000 updates
with batch size 64 and learning rate $10^{-4}$.
Stage 2 continues for 2{,}000 updates with batch
size 32 and lower learning rates:
$3\times10^{-5}$ for the correction tower and
key/value adapters, and $10^{-5}$ for LoRA.
Both stages use AdamW~\cite{loshchilov2017decoupled},
bf16 autocast, and data-parallel training on GH200 GPUs.

\paragraph{Data.}
\Cref{tab:supp_training_data} summarizes the training
sources and their exposure.
Syn4D~\cite{jiang2026syn4d} provides dense labels;
Dynamic Replica and PointOdyssey provide 3D tracks;
and hand--object datasets provide sparse annotations.
CO3D~\cite{reizenstein21co3d} and
ScanNet++~\cite{yeshwanth2023scannet++}
provide static-scene labels and are used only in stage 1.
Training windows are sampled with fixed dataset weights
and randomized temporal intervals.
Some windows include carried history from cached rollouts
to train the history-conditioning input.
Dataset splits are detailed in \cref{supp:exposure}.

\paragraph{Training augmentation.}
Each training window is sampled with a random temporal
stride. To improve robustness to coarse trajectory
quality, we perturb proposal motion while keeping
$\mathbf P(0)$ fixed.
On densely labeled windows, motion is retained
with probability 0.5, replaced by ground-truth
displacement with probability 0.1, or interpolated
with probability 0.4:
\[
\begin{aligned}
\widetilde{\mathbf P}(t)={}&\mathbf P(0)
+(1-\alpha)(\mathbf P(t)-\mathbf P(0))\\
&+\alpha(\mathbf X^\star(t)-\mathbf X^\star(0)),
\quad \alpha\sim\mathcal U(0,1).
\end{aligned}
\]
Independently, proposal motion is set to zero with
probability 0.3, and the encoded coarse-motion
channels are dropped with probability 0.5,
while projections and depths still follow the
perturbed proposal.
The static-input ablation (Tab.~\ref{tab:ablation_static}) therefore
evaluates a training-seen condition, while its
gap to full 4RC inputs shows the benefit of
coarse motion.
Targets are recomputed after augmentation.
All training proposals originate from 4RC;
no other tracker is used.

\begin{table}[t]
\centering
\footnotesize
\setlength{\tabcolsep}{4pt}
\caption{\textbf{Training data of the local refiner} (\cref{supp:training}). Windows: distinct
training data in units of the 48-frame window; Exposure: share of the training exposure of the listed sources over both stages.
Labels: dense, every point; tracks, ground-truth 3D tracks of points visible in the first frame; sparse, annotated hand and object
points; static, static-scene points from ground-truth depth and camera poses. For datasets that are also evaluated on, see \cref{supp:exposure}.}
\label{tab:supp_training_data}
\begin{tabular}{l l r r}
\toprule
Source & Labels & Windows & Exposure (\%) \\
\midrule
Syn4D~\cite{jiang2026syn4d} & dense & 14{,}868 & 53.9 \\
Dynamic Replica~\cite{karaev2023dynamicstereo} & tracks & 2{,}568 & 8.7 \\
PointOdyssey~\cite{zheng2023pointodyssey} & tracks & 2{,}491 & 8.4 \\
HOT3D~\cite{banerjee2025hot3d} & sparse & 4{,}018 & 10.9 \\
OakInk2~\cite{zhan2024oakink2} & sparse & 3{,}153 & 9.1 \\
DexYCB~\cite{chao2021dexycb} & sparse & 3{,}017 & 5.9 \\
TACO~\cite{liu2024taco} & sparse & 746 & 0.4 \\
CO3D~\cite{reizenstein21co3d} & static & 4{,}120 & 1.8 \\
ScanNet++~\cite{yeshwanth2023scannet++} & static & 1{,}104 & 0.9 \\
\midrule
Total & & 36{,}085 & 100.0 \\
\bottomrule
\end{tabular}
\end{table}

\subsection{Runtime}
\label{supp:runtime}

The runtime in Sec.~\ref{sec:experiment} is measured on one RTX~4090
with the towers in bf16. Fix4D takes 9.0\,s per
48-frame window (median), including autoencoder
encoding and decoding, one forward pass of both
towers, and trajectory composition.

\subsection{Transfer to other providers}
\label{supp:transfer}

Fix4D uses only RGB video and the provider's
trajectories, cameras, and validity masks.
To transfer across providers, we convert their
outputs into the same source-indexed representation
used for 4RC, without retraining.

\paragraph{Conditioning field.}
At frame 0, each source pixel is unprojected using
the provider's first-frame depth and camera rays.
At subsequent frames, displacements sampled from
the provider's tracks are bilinearly interpolated
onto the source grid.
All quantities remain in the provider's coordinate
system and scale, with its validity and depth
cutoffs respected.

\paragraph{Applying corrections.}
Fix4D predicts corrections from this conditioning
field, which are sampled at the query pixels and
added to the provider's own query trajectories.
Thus, the ``+ Fix4D'' results in Tabs.~\ref{tab:general}--\ref{tab:providers}
refine the original provider tracks rather than
replacing them with trajectories reconstructed
from the conditioning field.

\paragraph{Provider settings.}
SpatialTrackerV2~\cite{xiao2025spatialtrackerv2}
uses its released offline model, evaluated on the
benchmark queries and separately on a grid of every
sixth pixel for conditioning.
D4RT~\cite{zhang2026efficiently} uses the public
community reimplementation with its 48-frame
checkpoint and $256{\times}256$ inputs (Open-D4RT); its dense
trajectories are sampled every two pixels, with
cameras recovered by rigid alignment of its
predicted geometry.
V-DPM~\cite{sucar2026v} uses its official model,
dense first-view point maps sampled every two
pixels, and its estimated cameras and intrinsics.

\paragraph{Long videos.}
V-DPM is evaluated in 48-frame chunks, composed
as for 4RC (\cref{supp:handover}), with tracks
linked through the first shared frame.
An additional point-head evaluation per chunk
provides the geometry for chunk alignment.
Fix4D and its association tower are applied
without retraining.

\section{Handover refinement for long videos}
\label{supp:handover}

\paragraph{Chunked rollout.}
Following Point4D~\cite{jeon2026point4d}, long videos
are processed in 48-frame chunks with eight-frame
overlap (stride 40).
For 300-frame videos, chunks start at
$0,40,\ldots,240,252$; excess overlap in the final
shifted chunk is replaced rather than blended.
Each chunk is aligned to the first chunk's coordinate
frame using a similarity transform fitted to the
provider's point maps in shared frames, retaining the 85\% of
correspondences with the smallest residual.
Fix4D corrects geometry and motion in the first
chunk, and motion only in subsequent chunks.

\paragraph{Baseline composition.}
Let $\mathbf H_i(j)$ denote the carried trajectory
over the eight shared frames $j=0,\ldots,7$, and
$\mathbf R_i(t)$ the locally refined incoming track.
Following the reprojection chaining of
Point4D~\cite{jeon2026point4d}, each carried point
is linked to an incoming track using the shared
frame with the best depth agreement.
The incoming track is anchored at the last shared
frame and blended over the overlap:
\begin{equation}
\begin{aligned}
\mathbf A_i(t)&=\mathbf R_i(t)
    +\mathbf H_i(7)-\mathbf R_i(7),\\
\mathbf B_i(j)&=(1-\tfrac{j}{7})\mathbf H_i(j)
    +\tfrac{j}{7}\mathbf A_i(j),\quad j=0,\ldots,7,\\
\mathbf B_i(t)&=\mathbf A_i(t),\qquad t\geq8.
\end{aligned}
\label{eq:supp_anchor_blend}
\end{equation}
Without handover refinement (``local only'' in
Tab.~\ref{tab:long}), $\mathbf B$ is the output.

\paragraph{Association tower.}
A separately trained copy of the local refiner
(``Association Wan'' in \cref{fig:method_two_stage})
reads the incoming proposal $\mathbf P$, carried
history $\mathbf H$, RGB frames, and camera information
around each handover.
It predicts a corrected trajectory $\mathbf C$,
a merge gate, an abstention score, and a selection gate.
The tower acts only on predicted-moving points, whose
carried or incoming displacement exceeds
$\max(0.05,\,0.01|z|)$ in provider units.
These are partitioned into the re-association set
$\mathcal A$, the unlinked moving points of the main paper,
whose carried history moves and is valid in all eight
shared frames but whose link is unreliable
(outside the image or exceeding 5\% depth mismatch
at the first shared frame),
and the overlap-correction set $\mathcal M$ of the
remaining moving points with a finite tower prediction.
Other points retain $\mathbf B$.

\paragraph{Handover correction.}
For points in $\mathcal A$, candidate incoming tracks
are sampled on a $33\times33$ grid ($\pm64$\,px,
4\,px spacing) around the previous chunk's projected
estimate, with an additional abstention option.
An MLP scores candidates using geometric consistency
with $\mathbf C$ and $\mathbf H$.
A candidate is accepted if it passes the abstention
and depth-consistency checks; otherwise the baseline
composition is retained.
Accepted candidates replace the incoming tracks
before the local refinement pass.
The refined selected trajectory $\mathbf R^{\rm sel}$
is then merged using a selection gate
$g_i^{\rm sel}\in[0,1]$:
\begin{equation}
\begin{aligned}
\mathbf Y_i(j)&=\mathbf B_i(j)
 +g_i^{\rm sel}
 \bigl(\mathbf R_i^{\rm sel}(j)-\mathbf B_i(j)\bigr),\\
\mathbf Y_i(t)&=\mathbf R_i^{\rm sel}(t)
 +\mathbf Y_i(7)-\mathbf R_i^{\rm sel}(7),
 \quad t\geq8.
\end{aligned}
\label{eq:supp_selection_merge}
\end{equation}
For points in $\mathcal M$, the gated merge of
Eq.~(\ref{eq:learned_merge}) is applied over the shared frames.
Its final offset is propagated to subsequent frames:
\begin{equation}
\mathbf Y_i(t)=\mathbf B_i(t)
 +\mathbf Y_i(7)-\mathbf B_i(7),\quad t\geq8.
\label{eq:supp_future_propagation}
\end{equation}
This adjusts the placement of the new frames without
changing their frame-to-frame motion.

\paragraph{Training.}
The association tower is initialized from the
trained local refiner and trained separately on
handover pairs from PointOdyssey, Dynamic Replica,
Syn4D, and OakInk2, using simulated errors or
cached 4RC rollouts.
No other provider is used in training.
The objective combines trajectory, static-point,
seam, gate, and candidate-selection losses.
Training uses AdamW with batch size 16 for
approximately 1{,}200 updates.

\section{Evaluation protocol and benchmarks}
\label{supp:eval}

\begin{table}[t]
\centering
\footnotesize
\setlength{\tabcolsep}{5pt}
\caption{\textbf{Benchmark composition.} Clips, frames per clip, median annotated queries (including points that are not scored, \cref{supp:hoibench}) per clip, and share of scored query points that move (\%).}
\label{tab:supp_composition}
\begin{tabular}{l r r r r}
\toprule
Dataset & Clips & Frames & Queries & Moving (\%) \\
\midrule
\multicolumn{5}{l}{\emph{WorldTrack, 172 videos}} \\
ADT & 49 & 48 & 430 & 19.9 \\
DR & 50 & 48 & 830 & 24.9 \\
PO & 23 & 48 & 552 & 32.6 \\
PStudio & 50 & 48 & 160 & 95.3 \\
\midrule
\multicolumn{5}{l}{\emph{HOI benchmark, 401 clips}} \\
HO3D & 47 & 48 & 734 & 11.0 \\
H2O & 184 & 48 & 361 & 9.6 \\
DexYCB & 96 & 48 & 117 & 30.4 \\
ARCTIC & 31 & 48 & 108 & 86.5 \\
HOT3D & 7 & 48 & 421 & 21.4 \\
TACO & 36 & 48 & 225 & 72.5 \\
\midrule
\multicolumn{5}{l}{\emph{Stress test (not in the HOI average)}} \\
POV-Surgery & 54 & 48 & 438 & 40.6 \\
\midrule
\multicolumn{5}{l}{\emph{Long videos, first 300 frames}} \\
DR & 20 & 300 & 1766 & 33.2 \\
PO & 20 & 300 & 2000 & 70.0 \\
HO3D & 16 & 300 & 716 & 15.0 \\
\bottomrule
\end{tabular}
\end{table}

\subsection{Benchmarks}
\label{supp:benchmarks}
\Cref{tab:supp_composition} summarizes the evaluation sets. We weight datasets equally and report all-point and dynamic-point
scores separately, since the datasets differ strongly in size and in their share of moving points.

\paragraph{WorldTrack.}
We use the evaluation videos of ADT~\cite{pan2023aria}, Dynamic Replica~\cite{karaev2023dynamicstereo},
PointOdyssey~\cite{zheng2023pointodyssey} and Panoptic Studio~\cite{joo2015panoptic} of the WorldTrack
benchmark~\cite{feng2025st4rtrack}, with their images, ground-truth tracks and ground-truth cameras unchanged. As in WorldTrack, the
queries are points of the first frame and predictions are scored in the first camera frame; we score the first 48 frames of each
video, the clip length of D4RT~\cite{zhang2026efficiently} and of Fix4D. One of the 50 ADT videos is excluded because none of its tracks is visible in the first frame.
On WorldTrack, every query is scored.

\paragraph{Long videos.}
Following the long-video evaluation of Point4D~\cite{jeon2026point4d} on Dynamic Replica and PointOdyssey, the long-video set
uses the first 300 frames (the full length of a Dynamic Replica video) of 20 Dynamic Replica and 20 PointOdyssey videos from the official test splits of these datasets, and
adds 16 HO3D videos, each the first 300 annotated frames of a sequence. Here a point is dynamic if it moves by more than the threshold at any time within the 300 frames.
The single median scale is fitted over the whole 300 frames; the ``last 48'' columns apply the same scale and score only frames
252--299, which form the final chunk of the rollout (\cref{supp:handover}).

\paragraph{HOI benchmark.}
The HOI benchmark of the main paper contains 401 clips of 48 frames from six hand--object datasets; its construction is described in
\cref{supp:hoibench}. In addition, we report 54 POV-Surgery clips, built in the same way, as a separate stress test
(\cref{supp:pov}); they are not part of the HOI benchmark, of its average, or of Tab.~\ref{tab:providers}.

\begin{table*}[t]
\centering
\setlength{\tabcolsep}{3pt}
\caption{\textbf{Long-video 3D tracking, all points.} Companion to the main-paper long-video table (dynamic points): APD (\%, $\uparrow$) and EPE (m, $\downarrow$) on \emph{all} points over the first 300 frames and their final 48, on 20 DR, 20 PO and 16 HO3D videos, one median scale per sequence. Avg is the three-dataset mean. \first{Best}, \second{second}, and \third{third} are highlighted.}
\label{tab:supp_long_all}
\resizebox{0.9\textwidth}{!}{%
\begin{tabular}{lcccccccc}
\toprule
 & \multicolumn{2}{c}{DR} & \multicolumn{2}{c}{PO} & \multicolumn{2}{c}{HO3D} & \multicolumn{2}{c}{Avg} \\
\cmidrule(lr){2-3} \cmidrule(lr){4-5} \cmidrule(lr){6-7} \cmidrule(lr){8-9}
 & APD$\uparrow$ & EPE$\downarrow$ & APD$\uparrow$ & EPE$\downarrow$ & APD$\uparrow$ & EPE$\downarrow$ & APD$\uparrow$ & EPE$\downarrow$ \\
Method & Full / Last 48 & Full / Last 48 & Full / Last 48 & Full / Last 48 & Full / Last 48 & Full / Last 48 & Full / Last 48 & Full / Last 48 \\
\midrule
\multicolumn{9}{l}{\emph{Sparse tracking}} \\
TAPIP3D~\cite{zhang2026tapip3d} & 75.7 / 73.1 & 0.243 / 0.265 & 50.4 / 42.8 & 1.236 / 1.806 & 85.1 / 84.5 & \third{0.155} / 0.160 & 70.4 / 66.8 & 0.545 / 0.744 \\
\midrule
\multicolumn{9}{l}{\emph{Dense tracking}} \\
MonST3R~\cite{zhang2025monst3r} & 60.5 / 60.1 & 0.385 / 0.390 & 28.4 / 27.7 & 1.704 / 2.033 & 77.8 / 77.3 & 0.234 / 0.234 & 55.6 / 55.0 & 0.774 / 0.886 \\
St4RTrack~\cite{feng2025st4rtrack} & 76.6 / \third{75.8} & 0.229 / \third{0.235} & 35.4 / 33.2 & 1.105 / 1.491 & 77.9 / 78.4 & 0.203 / 0.197 & 63.3 / 62.5 & 0.512 / 0.641 \\
Any4D~\cite{karhade2026any4d} & \third{77.4} / 75.3 & \third{0.223} / 0.242 & 47.0 / 41.1 & 0.891 / 1.370 & \third{85.4} / \first{85.0} & \second{0.154} / 0.156 & 69.9 / 67.1 & 0.423 / 0.589 \\
TrackCraft3R~\cite{nam2026trackcraft3r} & 71.3 / 71.0 & 0.288 / 0.293 & 44.8 / 43.0 & 1.287 / 1.674 & 84.3 / 84.3 & \third{0.155} / \third{0.154} & 66.8 / 66.1 & 0.577 / 0.707 \\
Point4D~\cite{jeon2026point4d} & 73.3 / 69.4 & 0.271 / 0.311 & 46.1 / 36.6 & 1.103 / 1.662 & 82.3 / 80.7 & 0.174 / 0.179 & 67.2 / 62.2 & 0.516 / 0.717 \\
V-DPM~\cite{sucar2026v} & 67.6 / 65.4 & 0.296 / 0.318 & 57.2 / 49.2 & \third{0.695} / \third{1.173} & 84.1 / 82.9 & 0.161 / 0.169 & 69.6 / 65.8 & 0.384 / 0.553 \\
4RC~\cite{luo20264rc} & 76.7 / 74.0 & 0.226 / 0.251 & 55.6 / 48.8 & 0.878 / 1.353 & 83.6 / 81.9 & 0.173 / 0.185 & \third{72.0} / \third{68.2} & 0.426 / 0.596 \\
\midrule
\multicolumn{9}{l}{\emph{Fix4D}} \\
4RC~\cite{luo20264rc} + Fix4D (local only) & \second{85.1} / \second{81.9} & \second{0.156} / \second{0.182} & 57.4 / 50.0 & 0.704 / 1.191 & 84.4 / 83.3 & 0.159 / 0.165 & \second{75.6} / \second{71.7} & \second{0.340} / \second{0.513} \\
4RC~\cite{luo20264rc} + Fix4D (w.\ handover refinement) & \first{85.5} / \first{82.7} & \first{0.152} / \first{0.175} & \third{57.6} / \third{50.1} & 0.699 / 1.187 & 84.4 / 83.4 & 0.159 / 0.163 & \first{75.8} / \first{72.1} & \first{0.337} / \first{0.509} \\
V-DPM~\cite{sucar2026v} + Fix4D (local only) & 68.8 / 66.6 & 0.280 / 0.302 & \second{58.3} / \second{50.3} & \second{0.680} / \second{1.156} & \second{85.6} / \third{84.7} & \first{0.144} / \second{0.150} & 70.9 / 67.2 & 0.368 / 0.536 \\
V-DPM~\cite{sucar2026v} + Fix4D (w.\ handover refinement) & 68.8 / 66.7 & 0.280 / 0.301 & \first{58.6} / \first{50.4} & \first{0.672} / \first{1.146} & \first{85.7} / \second{84.9} & \first{0.144} / \first{0.148} & 71.0 / 67.3 & \third{0.365} / \third{0.532} \\
\bottomrule
\end{tabular}}
\end{table*}

\begin{figure*}[t]
\centering
\includegraphics[width=0.85\linewidth]{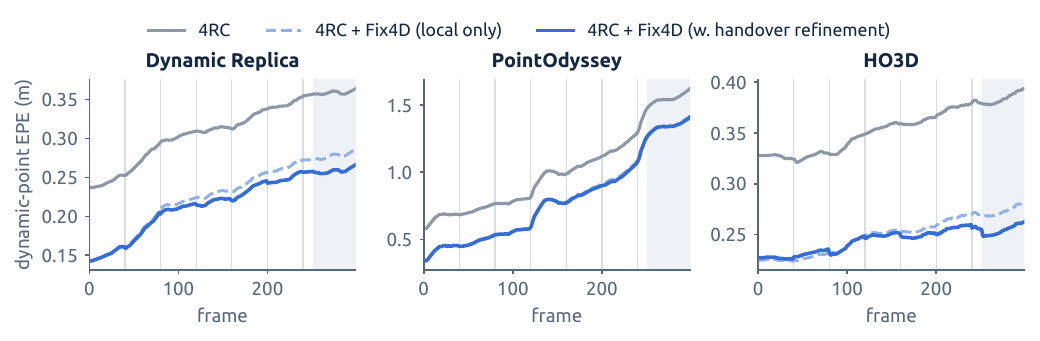}
\caption{\textbf{Error over time on long videos.} Per-frame dynamic-point EPE over the 300 frames of the long-video set
(Tab.~\ref{tab:long}), averaged over the videos of each dataset, for 4RC (chained 48-frame chunks) and 4RC + Fix4D without and with
handover refinement.
Shaded: the last 48 frames (the final chunk).}
\label{fig:supp_long_curves}
\end{figure*}
\begin{table}[t]
\centering
\setlength{\tabcolsep}{3pt}
\caption{\textbf{POV-Surgery: off-center principal point stress test.} All methods on the 54 POV-Surgery clips (48 frames; 49 clips
have moving points). The camera's principal point lies far from the image center ($c_y = 175$ of 1{,}080 rows). All methods reach 46--53\% APD on all points. APD (\%, higher is
better) and EPE (m, lower is better) on all and dynamic points, one median scale per sequence. $\Delta$ Dyn.\ EPE: change of the
dynamic-point EPE of a Fix4D row against its own provider row, computed from unrounded scores. We highlight the \first{best}, \second{second} and \third{third}
results.}
\label{tab:supp_pov}
\resizebox{\linewidth}{!}{%
\begin{tabular}{lccc}
\toprule
 & APD$\uparrow$ & EPE$\downarrow$ & $\Delta$ Dyn.\ EPE \\
Method & All / Dyn. & All / Dyn. & vs.\ provider \\
\midrule
\multicolumn{4}{l}{\emph{Sparse tracking}} \\
STv2~\cite{xiao2025spatialtrackerv2} & 52.6 / 64.8 & \second{0.443} / 0.306 &  \\
TAPIP3D~\cite{zhang2026tapip3d} & 49.9 / 60.5 & 0.479 / 0.333 &  \\
\midrule
\multicolumn{4}{l}{\emph{Dense tracking}} \\
MonST3R~\cite{zhang2025monst3r} & 48.2 / 59.5 & 0.500 / 0.340 &  \\
St4RTrack~\cite{feng2025st4rtrack} & 47.3 / 60.7 & 0.525 / 0.344 &  \\
Any4D~\cite{karhade2026any4d} & 52.3 / 63.1 & 0.446 / 0.314 &  \\
TraceAnything~\cite{liu2026trace} & 46.4 / 56.5 & 0.566 / 0.445 &  \\
TrackCraft3R~\cite{nam2026trackcraft3r} & 51.9 / 64.9 & 0.459 / 0.297 &  \\
Point4D~\cite{jeon2026point4d} & 51.9 / 64.4 & 0.456 / 0.305 &  \\
V-DPM~\cite{sucar2026v} & \third{52.8} / 66.3 & 0.452 / 0.291 &  \\
D4RT~\cite{zhang2026efficiently} & 51.8 / 64.6 & 0.450 / 0.306 &  \\
4RC~\cite{luo20264rc} & 51.3 / 62.0 & 0.461 / 0.322 &  \\
\midrule
\multicolumn{4}{l}{\emph{Fix4D}} \\
4RC~\cite{luo20264rc} + Fix4D & 52.5 / \third{66.4} & 0.447 / \third{0.282} & $-4.0$\,cm ($-13\%$) \\
STv2~\cite{xiao2025spatialtrackerv2} + Fix4D & \second{53.2} / \second{67.9} & \first{0.435} / \first{0.269} & $-3.7$\,cm ($-12\%$) \\
D4RT~\cite{zhang2026efficiently} + Fix4D & 51.7 / 65.8 & \third{0.445} / 0.284 & $-2.2$\,cm ($-7\%$) \\
V-DPM~\cite{sucar2026v} + Fix4D & \first{53.3} / \first{68.6} & 0.447 / \second{0.274} & $-1.7$\,cm ($-6\%$) \\
\bottomrule
\end{tabular}
}
\end{table}

\begin{figure*}[t]
\centering
\includegraphics[width=0.88\linewidth]{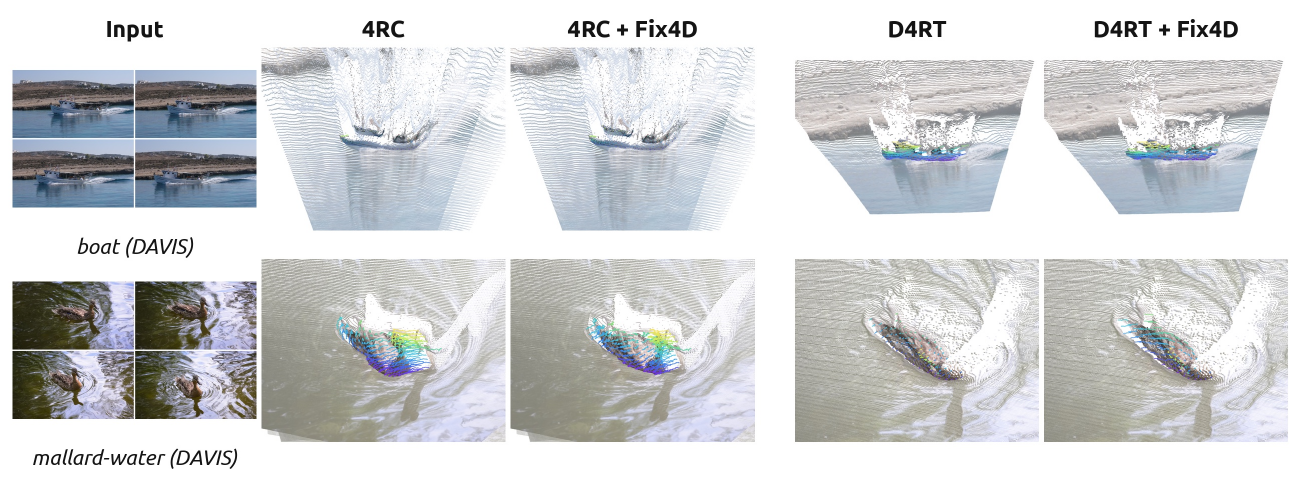}
\caption{\textbf{Failure cases.}
DAVIS~\cite{perazzi2016benchmark} clips of a boat (top)
and a swimming duck (bottom): input frames (left)
and trajectories from 4RC~\cite{luo20264rc} and
D4RT~\cite{zhang2026efficiently}, without and with Fix4D.
Fix4D fails to recover the object motion missed by
the initial trackers (Sec.~\ref{supp:limits}), under challenging coupled camera, foreground motion and background ambiguities.}
\label{fig:supp_failure}
\end{figure*}

\subsection{Metrics and baselines}
\label{supp:metric}
We follow the 3D trajectory evaluation protocols of Point4D~\cite{jeon2026point4d}, St4RTrack~\cite{feng2025st4rtrack} and
Any4D~\cite{karhade2026any4d}. Predictions are compared with the ground truth in the first camera frame after alignment with one
median-norm scale per evaluated sequence (48 or 300 frames; the median ground-truth norm divided by the median predicted norm); no rotation or translation is fitted.
EPE is the mean end-point error in meters, and APD is the percentage of point--frames within $\delta$ of the ground truth, averaged
over $\delta\in\{0.1,\,0.3,\,0.5,\,1.0\}$\,m. Occluded point--frames are included, since the task is 3D tracking through occlusion.
The dynamic-point columns average over moving points only, with the same scale. The dynamic-point criterion is our own and is applied
identically to every method: a point is moving if its largest ground-truth displacement from the first frame exceeds
$\max(5\,\mathrm{cm},\,0.01\,z_0)$, where $z_0$ is its first-frame depth.
A dataset's score is the mean over its clips (clips without moving points do not enter the dynamic-point columns), and ``Avg'' is the
mean over datasets.

\paragraph{Baselines.}
All baselines use their released code and weights (for D4RT, the public reimplementation) with their own depth and camera
estimates, and are scored with this protocol on the same clips and queries. MonST3R is run pairwise through St4RTrack's inference
code. On long videos, Any4D and V-DPM are run per 48-frame chunk and composed as for 4RC, Point4D uses its released long-video
mode, and TrackCraft3R computes its geometry in streaming mode with 60-frame chunks.

\subsection{Construction of the HOI benchmark}
\label{supp:hoibench}
The clip format and the queries follow WorldTrack~\cite{feng2025st4rtrack}, which builds on TAPVid-3D~\cite{koppula2024tapvid}:
each clip provides the RGB frames, ground-truth intrinsics, query points in the first frame, and the ground-truth 3D position
of every query point in every frame, in meters in the first camera frame, with per-frame visibility.
The ground-truth tracks follow TAPVid-3D and D4RT~\cite{zhang2026efficiently}: a track is a physical point transported by the
annotated poses. Object points are mesh vertices under the annotated 6-DoF object pose, hand points follow the posed hand model
(MANO~\cite{MANO:SIGGRAPHASIA:2017} or the dataset's own hand model), and static background points are lifted from the dataset's depth maps
(for HOT3D, taken from the scene's SLAM point cloud) and held fixed in the world.
The benchmark uses the annotated sequences of HO3D~v3~\cite{hampali2020honnotate} (evaluation only,
\cref{supp:exposure}), subjects 1--4 of H2O~\cite{kwon2021h2o}, subject 01 of
DexYCB~\cite{chao2021dexycb}, subject s04 of ARCTIC~\cite{fan2023arctic}, 7 captures of HOT3D~\cite{banerjee2025hot3d}, and 36 captures
of TACO~\cite{liu2024taco}; the additional POV-Surgery clips~\cite{wang2023pov} are built in the same way. Clip counts are in \cref{tab:supp_composition}.
Hand, object and background points are scored, except the object points of H2O (bounding-box keypoints) and the hands of TACO
(image-based fits).

Recordings with invalid labels are dropped, using checks on the annotations and the sensor data only (such as the agreement between
sensor depth and the annotated geometry), and a time and rotation offset between labels and video on TACO is corrected from the labels
and the images alone; no check uses the output of a method.
On every dataset, we verified that the labels agree with the image content: the correlation of image patches along the
ground-truth tracks is highest at the labeled timing.

\subsection{Training exposure}
\label{supp:exposure}

No evaluation clip is used to train Fix4D.
\emph{Held-out datasets.}
The local refiner is not trained on HO3D, H2O, ARCTIC,
POV-Surgery, ADT, or Panoptic Studio.
\emph{Held-out sequences.}
For DexYCB, HOT3D, and TACO, training and evaluation
use disjoint captures: we train on DexYCB subjects
03--10 and evaluate on subject 01; none of the HOT3D evaluation captures is among the training captures;
and none of the TACO evaluation captures is used
for training.
Dynamic Replica and PointOdyssey are commonly used
for training 4D reconstruction and tracking methods,
including 4RC~\cite{luo20264rc},
D4RT~\cite{zhang2026efficiently}, and
Point4D~\cite{jeon2026point4d}.
We likewise use only their official training splits,
with no overlap with WorldTrack evaluation scenes
or the official test videos used for long-video evaluation.
\emph{Association tower.}
It is trained on Syn4D and OakInk2 handovers and on
handover pairs from the official training splits of
Dynamic Replica and PointOdyssey, without HO3D data.
The held-out distinctions refer to the training data
of both Fix4D components (local refiner and association
tower); the providers and video-model initialization
weights are used as released.

\section{Additional results}
\label{supp:results}

\subsection{Long videos}
\label{supp:long}

The 300-frame long-video set includes Dynamic Replica,
PointOdyssey, and HO3D (\cref{supp:benchmarks}), excluding
Panoptic Studio (150-frame videos) and egocentric datasets,
where most first-frame points leave the field of view.
\Cref{tab:supp_long_all} complements Tab.~\ref{tab:long} with all-point
results: 4RC + Fix4D achieves the lowest average EPE
over full videos and their last 48 frames
(0.337 / 0.509\,m), while V-DPM + Fix4D achieves
the lowest EPE on PointOdyssey and HO3D.
Handover refinement further reduces dynamic-point EPE
for both providers, with larger gains in the last
48 frames (Tab.~\ref{tab:long} of the main paper), particularly on Dynamic Replica
and HO3D (\cref{fig:supp_long_curves}).

\subsection{POV-Surgery stress test}
\label{supp:pov}

POV-Surgery~\cite{wang2023pov} tests robustness
to an off-center principal point
($c_y=175$ of $1{,}080$ rows).
All methods achieve only 46--53\% all-point APD
(\cref{tab:supp_pov}).
Nevertheless, Fix4D reduces the dynamic-point EPE
of all four providers by 1.7--4.0\,cm (6--13\%).

\section{Failure cases and limitations}
\label{supp:limits}

Fix4D refines existing trajectories without updating
the provider's camera estimates, so it may not fully
recover from severe reconstruction errors.
On POV-Surgery, for example, all refined methods
remain below 54\% all-point APD (\cref{supp:pov}).
\Cref{fig:supp_failure} shows two DAVIS failure cases where 4RC
and D4RT predict little object motion, which Fix4D
does not recover.
In the boat sequence, 4RC's estimated camera motion
keeps nearly static 3D tracks projected onto the
moving boat, illustrating an ambiguity between
camera and object motion.

For long videos, Fix4D builds on the provider's
chunked alignment and trajectory composition
(\cref{supp:handover}).
Although handover refinement brings larger gains
in the last 48 frames, its contribution remains
smaller than that of local refinement (Tab.~\ref{tab:long}).
The association tower observes only the eight shared
frames and the first new frame, and earlier frames
are not revisited.
While overlap corrections are predicted in 3D,
re-association still selects among incoming tracks
near the carried point's image projection.
Occluded or out-of-view points may therefore lack
valid continuation candidates, leaving the tower
to abstain.
Recovering correspondences beyond the overlap,
such as after reappearance, remains future work.

\end{document}